\documentclass[journal, twoside]{IEEEtran}

\usepackage{graphicx}
\usepackage{wrapfig}
\usepackage{colortbl}
\usepackage{float}
\usepackage{overpic}
\usepackage{multirow}
\usepackage{booktabs}
\usepackage{bm}
\usepackage{verbatim}
\usepackage{color}
\usepackage{arydshln}
\usepackage{makecell,rotating}
\usepackage{array}
\usepackage{amsmath}
\usepackage{amssymb}
\usepackage{amsthm}
\usepackage{graphicx}
\usepackage{subfigure}
\usepackage{cite}
\usepackage{color}
\usepackage{epstopdf}
\usepackage{subeqnarray}
\usepackage{cases}
\usepackage{mathrsfs}
\usepackage{multirow}
\usepackage{algorithm}
\usepackage{algorithmicx}
\usepackage{algpseudocode}
\floatname{algorithm}{Algorithm}

\usepackage[dvipsnames]{xcolor}
\usepackage{enumitem}

\definecolor{citecolor}{RGB}{119,185,0}
\definecolor{citecolor1}{RGB}{66,168,235}
\usepackage{hyperref}
\hypersetup{colorlinks=true,citecolor=citecolor1,linkcolor=BrickRed,urlcolor=Thistle}

\definecolor{mygray}{gray}{.88}
\definecolor{mygrayd}{gray}{.94}

\def\ie{\emph{i.e.}}
\def\eg{\emph{e.g. }}
\def\etc{\emph{etc}}
\def\etal{{\em et al.~}}

\begin{document}

\title{Phrase-Based Affordance Detection \\ via Cyclic Bilateral Interaction}

\author{Liangsheng Lu$^{\textbf{*}}$, Wei Zhai$^{\textbf{*}}$, Hongchen Luo, Yu Kang, 
\IEEEmembership{Senior Member, IEEE} and Yang Cao, \IEEEmembership{Member, IEEE}

\thanks{$\textbf{*}$ Liangsheng Lu and Wei Zhai contributed equally to this paper.}

\thanks{Liangsheng Lu, Wei Zhai, Hongchen Luo, Yu Kang, Yang Cao are at the University of Science and Technology of China, Anhui, China. (email: \{lulsheng, wzhai056, lhc12\}@mail.ustc.edu.cn, \{kangduyu, forrest\}@ustc.edu.cn).}
}

\maketitle

\begin{abstract}
Affordance detection, which refers to perceiving objects with potential action possibilities in images, is a challenging task since the possible affordance depends on the person’s purpose in real-world application scenarios. The existing works mainly extract the inherent human-object dependencies from image/video to accommodate affordance properties that change dynamically. In this paper, we explore to perceive affordance from a vision-language perspective, and consider the challenging phrase-based affordance detection problem, i.e., given a set of phrases describing the action purposes, all the object regions in a scene with the same affordance should be detected. To this end, we propose a cyclic bilateral consistency enhancement network (CBCE-Net) to align language and vision features progressively. Specifically, the presented CBCE-Net consists of a mutual guided vision-language module that updates the common features of vision and language in a progressive manner, and a cyclic interaction module (CIM) that facilitates the perception of possible interaction with objects in a cyclic manner. In addition, we extend the public Purpose-driven Affordance Dataset (PAD) by annotating affordance categories with short phrases. The contrastive experimental results demonstrate the superiority of our method over nine typical methods from four relevant fields in terms of both objective metrics and visual quality. The related code and dataset will be released at \url{https://github.com/lulsheng/CBCE-Net}.
\end{abstract}

\begin{IEEEImpStatement}
Affordance learning is concerned with the possible set of actions that an environment can offer to an actor. At present, almost all work focus on utilizing cues of human-object interactions in visual media such as images and videos to learn affordances. Few work explores affordance using language information, which is crucial to building more intelligent agents because we live a multimodal world. In this paper, we propose a phrase-based affordance detection task combining vision and language modalities. We first build a vision-language dataset based on the previous dataset. Then we propose a method to fuse and align multi-modal features. Our method outperforms relevant methods from several other fields. The proposed dataset and method can facilitate building more intelligent agents that can better comprehend the surrounding environment. For instance, the phases-based affordance detection can be used in building more intelligent home service systems. The agents can understand instructions of instructions without specific entities.

\end{IEEEImpStatement}


\section{Introduction}

\IEEEPARstart{T}{he} term ``Affordance'' is used to describe the interactions between humans, animal, and their environment. In other words, affordance implies the complementary between the animal and the environment \cite{gibson1977theory}. The affordance is regarded as an inherent property of an object independent of the user’s characteristics. Thus, investigating the affordance of objects leads agents to interact with environments better. In the field of vision, computer vision techniques not only need abilities to detect the content in a scene but also require the capability to infer possible interactions between humans, animals, and the corresponding environments \cite{gibson2014ecological}. Recently, affordance has drawn remarkable attention and has been widely explored in various application fields. For instance, the theory of affordance is applied to design more intelligent and more robust robotic systems in complex and dynamic environments \cite{horton2012affordances}. As a result,  perceiving the affordance of objects has a broad range of applications in several fields such as action recognition \cite{qi2017predicting,earley1970efficient,li2021tri}, scene parsing \cite{bagautdinov2017social, fang2018demo2vec,zhu2020self} and robot grasping \cite{yamanobe2017brief,shiraki2014modeling}, \etc. \par

\begin{figure}[t]
	\centering
	\begin{overpic}[width=1.\linewidth]{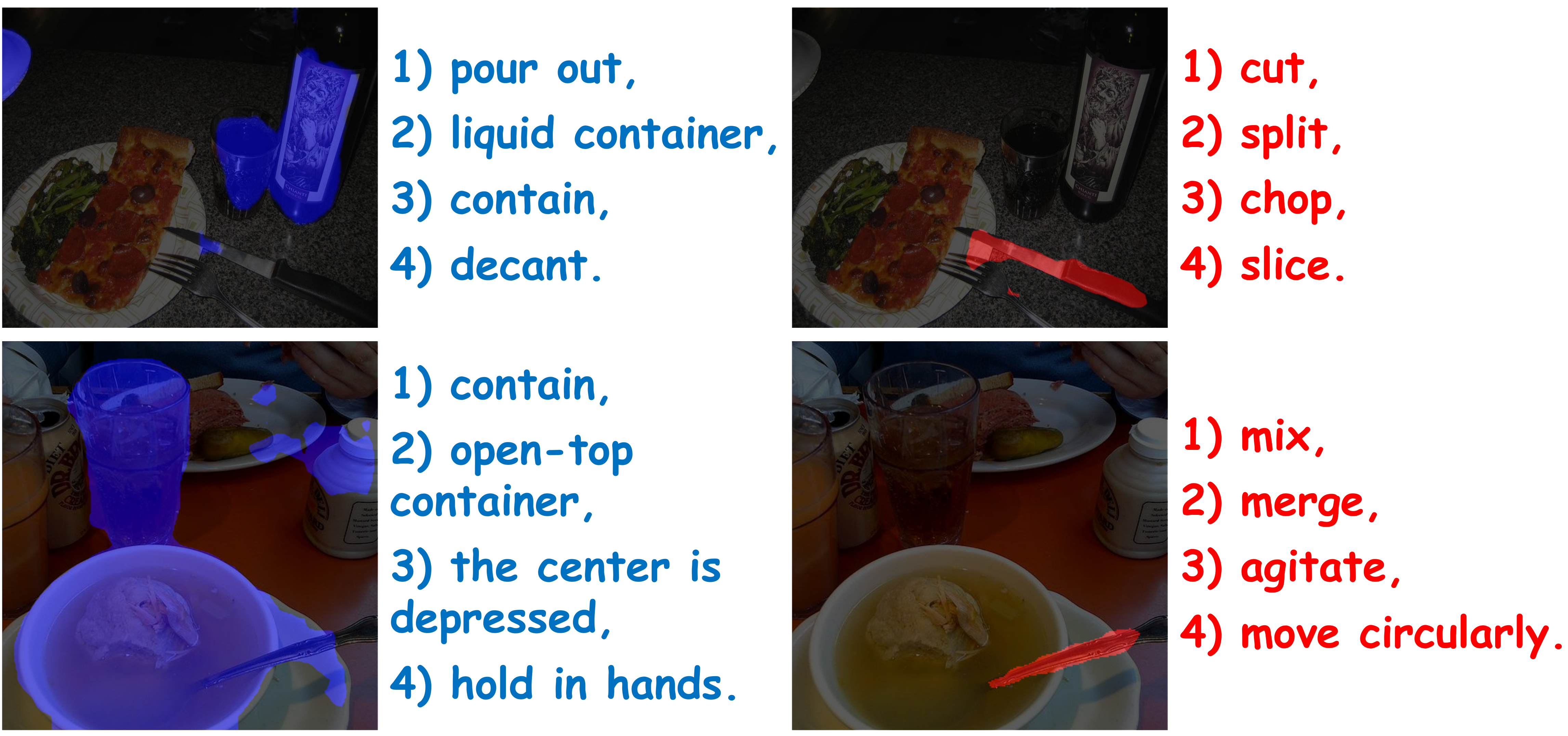}
		\end{overpic}
		\caption{\textbf{Illustration of perceiving affordance.} Given a set of phrases that describes the affordance, the corresponding objects could be detected. In the first row, the phrases in blue and red indicate affordance ``\textit{contain}''  and ``\textit{cut}'', then the corresponding objects ``\textit{beer bottle}'' and ``\textit{knife}'' with these affordances are segmented in blue and red, respectively. In the second row, the phrases in blue and red indicate affordances ``\textit{contain}'' and ``\textit{scoop}'', and the objects ``\textit{cup}'', ``\textit{bowl}'' and ``\textit{spoon}'' are highlighted in blue and red, respectively.}
	\label{FIG:fig_1}
\end{figure}

Previous work mainly addressed applications of affordance in the vision-only field. Some work \cite{sawatzky2017adaptive,zhao2020object,Oneluo} construct mapping relationships between objects representations and affordance categories. However, affordance is closely related to the environment and the actor, limiting models' generalization capability in new unseen scenes, leading to incorrect perception and localization. To solve the above problem, some other work perceives affordance objects by mining human-object interaction cues from videos or images \cite{Oneluo,kjellstrom2011visual,fang2018demo2vec,zhai2021one,luo2021learning} and transferring them to target images which enables models to better cope with the effects of dynamic changes of affordance and keep well generalization ability in new unseen environments. Unlike the above work, this paper attempts to explore the potential of language pairs in affordance detection tasks from a multimodal perspective. We consider utilizing a series of phrases combinations to describe affordances of an object and then generate segmentation masks from the corresponding images. This process is also consistent with the application scenarios where real intelligent agents receive information in multiple modalities from different sources to jointly perceive 
the affordances of objects. \par
\begin{figure*}[t]
	\centering
		\begin{overpic}[width=1.\linewidth]{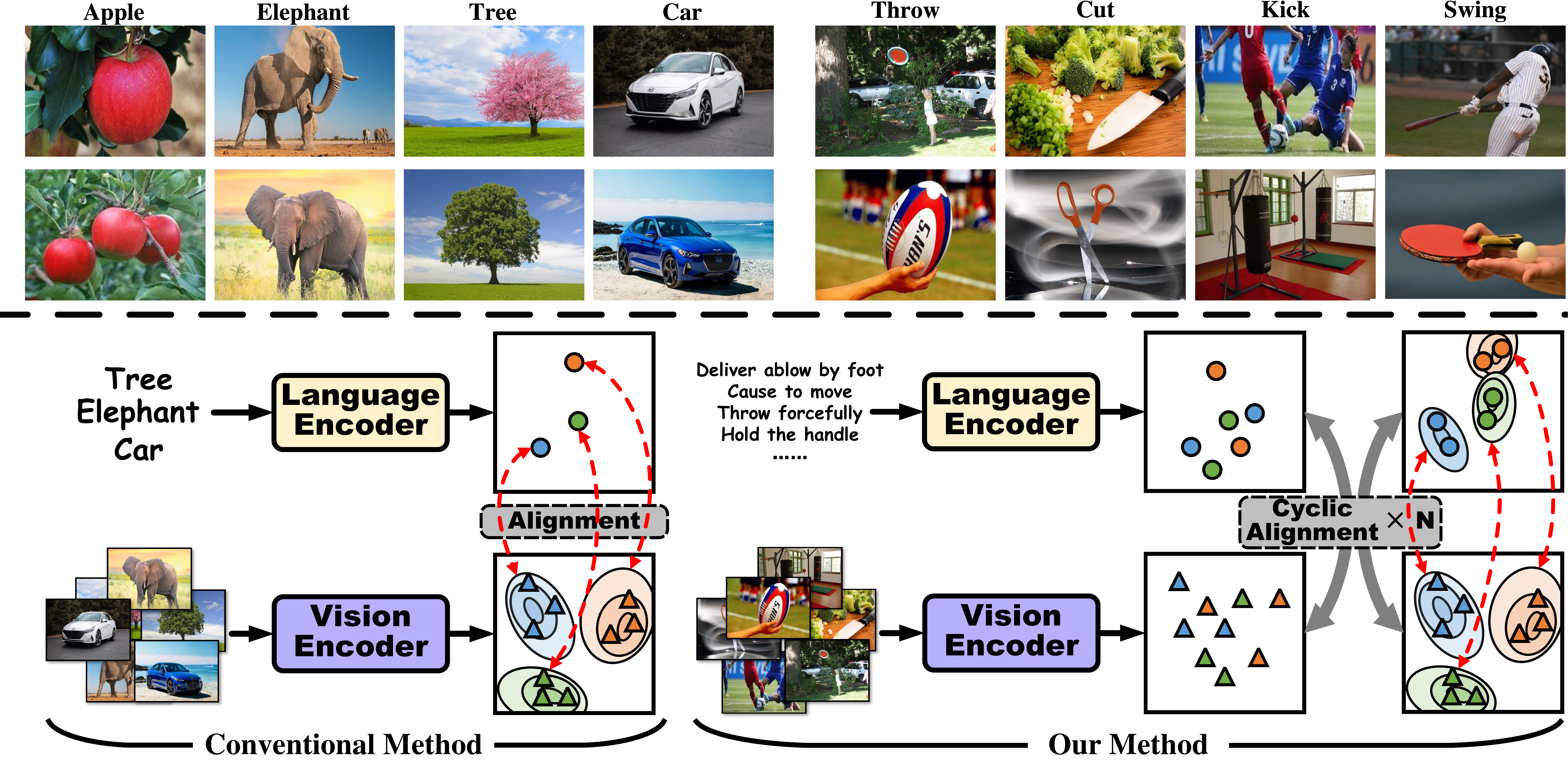}
		\put(-1,15){\textbf{(b)}}
		\put(-1,37.7){\textbf{(a)}}
		\end{overpic}
		\caption{\textbf{Task and method differences between affordance-related vision-language task and traditional ones.} \textbf{(a)} shows the problems caused by the multiplicity property of affordance. In traditional V-L tasks, the appearances of objects with the same language descriptions are generally similar, while the differences are significant in affordance-related V-L task. For images on the left, objects referred to by the same entity phrases are similar in color, shape, and texture; nevertheless, the opposite is true for images on the right. In \textbf{(b)},  We compare our method with conventional methods. In the traditional method, vision features are close enough in the distance in feature space, leading to easier alignment. However, for affordance-related V-L task, vision and language features are cluttered in feature space. We design a cyclic and bilateral mechanism to cope with these problems by enhancing iter-modal semantic consistency step by step. (See Section \ref{method} for details.)}
	\label{FIG:cyclic}
\end{figure*}
Therefore, we propose a phrase-based affordance detection task in this paper. That is, input a set of textual phrases describing affordances and an image, then the objects that can afford the corresponding affordance are expected to be segmented precisely (as shown in Fig. \ref{FIG:fig_1}). We choose natural language phrases to describe affordances without considering the specific object categories suitable in practical application scenarios. In practice, humans often communicate with each other with incomplete sentences and pass out some common-sense information that does not need to be explicitly pointed out. When a human interacts with an agent, likely, the given instructions are also incomplete \cite{chen2020enabling}. For example, a human asks the agent to ``\textit{pour me some water}'' while whether to use a cup or a bowl to pour the water is not indicated explicitly. This shows that learning common-sense knowledge such as affordance will lead to more intelligent agents. \par
Nevertheless, affordance is a special property different from the semantic category. One object may have multiple affordances, while one affordance may correspond to different objects. For example, the affordance of \textit{``Bed''} includes two different actions: \textit{``Sit''} and \textit{``Lie''}, while \textit{``Chair''} can also afford action \textit{``Sit''}. This may lead to great differences in the visual representation of different objects referred to by the same textual descriptions. Fig. \ref{FIG:cyclic} (a) shows the differences between traditional vision-language tasks and affordance-related V-L tasks. 
Such variations in the color, texture, and shape of objects would render affordance-related vision-language task more difficult in the alignment of textual and visual features compared to conventional multimodal tasks \cite{hu2016segmentation,liu2017recurrent,li2018referring,hu2020bi,jing2021locate}. 
The differences can lead to significant divergence in the distribution of textual and visual representations in feature space. It may be difficult for the deep-learning-based network to align features from these two modalities through a single learning step. Because the distribution of corresponding visual features seems to be irregular for the same textual representation, the cross-modal semantic consistency is difficult to capture if the network only updates features once a time. To tackle this problem, as shown in Fig. \ref{FIG:cyclic} (b), we design a cyclic bilateral update mechanism. Our model updates visual and linguistic features with the leverage of another modality to enhance the inter-modal semantic consistency step by step in a bilateral and cyclic manner. The inter-modal consistency is gradually enhanced after several cyclic alignments. \par
To this end, we propose a \textbf{C}yclic \textbf{B}ilateral \textbf{C}onsistency \textbf{E}nhancement Network (\textbf{CBCE-Net}), which consists of three main modules: \textbf{V}ision guide \textbf{L}anguage \textbf{M}odule (VLM), \textbf{L}anguage guide \textbf{V}ision \textbf{M}odule (LVM) and \textbf{C}yclic \textbf{I}nteraction \textbf{M}odule (CIM). Utilizing the attention mechanism, VLM learns the importance of linguistic features in each visual region and derives new linguistic features. Then LVM uses the output of the textual feature from VLM with aggregating multi-level information to guide the generation of new visual features. VLM and LVM operations are repeated several times in CIM module to enhance the inter-modal semantic consistency in a cyclic and bilateral manner. \par

The current affordance datasets lack explicit descriptions of affordance using natural language. To address this issue, based on the previously proposed PAD dataset \cite{Oneluo}, we annotate associated short phrases according to affordance categories, as this is more suitable for practical application scenarios. With the leverage of WordNet \cite{miller1995wordnet}, which is a hierarchical lexical database, we annotate the text of affordance from four different but closely related perspectives: potential actions, functions, appearance features, and the environment. We name the resulting dataset based on PAD with annotating natural language as PAD-Language dataset (PAD-L). \par
In summary, our contributions are four-folds:
\begin{itemize}[leftmargin=12pt]
    \item [1)]
    We propose a new task for object affordance detection based on text phrases. Inputting a set of text phrases and an image containing related objects, the corresponding segmentation masks are expected to be generated. This makes more intelligent agents better comprehend humans' intentions during interactions with humans and locate specific objects in the scene even if the instructions do not indicate the specific category of the object.
    \item [2)]
    We design a novel CBCE-Net to effectively extract the affordance information from the given set of text phrases and then segment the corresponding object regions in the given image. Our model can effectively solves the vision-language alignment issue caused by the multiplicity feature of affordance. The text and vision information could interact and align well with each other in a cyclic and bilateral manner even though the visual appearance, texture and color are highly diverse. 
    \item [3)] 
    We annotate affordance categories using natural language phrases based on the existing dataset PAD. A new affordance dataset with natural language descriptions PAD-L is constructed, which extends the affordance of objects from limited categories to unconstrained natural language. The new dataset can be used in various downstream tasks.
    \item [4)]
    Compared with nine different approaches chosen from four relevant fields, our model achieves the best result in both subjective and objective terms, which is able to serve as a strong baseline for future work.
\end{itemize}

The rest of the paper is organized as following. Section \ref{related} illustrates the previous work related to phrase based affordance detection task. Section \ref{anno} describes details to annotate text phrases. The novel proposed  CBCE-Net is introduced in Section \ref{method}. Section \ref{exp} shows the experiment results and analysis on PAD-L. We conclude the paper with discussing possible applications and future work in Section \ref{conclude}.

\section{Related Work} \label{related}

\subsection{Affordance Learning}

Visual Affordance has been extensively studied in computer vision and robotics communities because of its close association with action recognition, scene parsing, and human-robot interaction. Many approaches have been proposed to learn the visual affordance of objects in images. Hassan \etal \cite{hassan2015attribute} proposed a Bayesian network-based affordance detection method that exploits the attribute of the object, actor, and environment. Grabner \etal \cite{grabner2011makes} utilized a human skeleton 3D model to learn the action of sitting on a chair to infer whether an object can afford ``\textit{sitting}'' action or not. \par
With the development of deep neural networks, many methods based on deep learning have been proposed. Inspired by semantic segmentation \cite{long2015fully} approaches, affordance learning is extensively studied at the pixel level. Sawatzky \etal \cite{sawatzky2017adaptive} proposed a weakly supervised affordance segmentation method to predict the fine segmentation masks by effectively leveraging the weakly labeled data, which is annotated in image-level and key-points level. Nguyen \etal \cite{nguyen2017object} considered affordance segmentation as an object detection task. They employ the existing object detection models to obtain a set of candidate bounding box proposals. Afterward, atrous convolution is used to generate the final fine mask. Zhao \etal \cite{zhao2020object} proposed an end-to-end model to exploit the symbiotic relationship between multiple affordances with the combinational relationship between affordance and objectness to produce a pixel-level affordance map. \par
In addition to using the features of objects themselves, some recent work has leveraged auxiliary information to learn visual affordance \cite{thermos2017deep,wang2017transferring}.  Fang \etal \cite{fang2018demo2vec} proposed a method to learn the affordance of unseen objects from expert demonstration videos. Their model extracts feature embedding vectors from demonstration videos to predict the same objects' interaction regions and action labels in a given image. Luo \etal \cite{Oneluo} proposed an one-shot detection method to detect affordance in unseen scenarios. Their model first extracts intention information from support images. Then, the intention is transferred to query images to detect objects capable of affording the intention. To this end, they construct a new \textbf{P}urpose-driven \textbf{A}ffordance \textbf{D}ataset (PAD), which compensates for the lack of rich scenes in the previous datasets. \par
Unlike all the work mentioned above, where affordance is explored only in visual mediums, we attempt to investigate affordance detection involving natural language. Inputting a set of phrases that describe affordances and an image, the corresponding objects are expected to be segmented, which meets the realistic scenarios where robots receive information in multiple modalities from multiple sources.

\subsection{Referring Expression Grounding}
Given a piece of text, referring expression grounding task aims to comprehend the natural language content with locating the corresponding regions in the input image. Many efforts achieve localization at bounding box level \cite{mao2016generation,yu2017joint,liu2019improving}. Yu \etal \cite{yu2018mattnet} proposed a modular network which decomposes the input natural language description into subject, location, and relationship attributes to improve the localization performance. Liu \etal \cite{liu2021cross} adopted graph models with an attention mechanism to capture the relationship between the object regions in the given image. In association with visual affordance, Mi \etal \cite{mi2019object,mi2020intention} investigated the use of natural language to guide visual affordance detection. Their model first extracts the intention in the natural language then locates referred objects in the given image at the bounding box level. \par

\begin{figure*}[t]
\centering
  \includegraphics[width=0.99\textwidth]{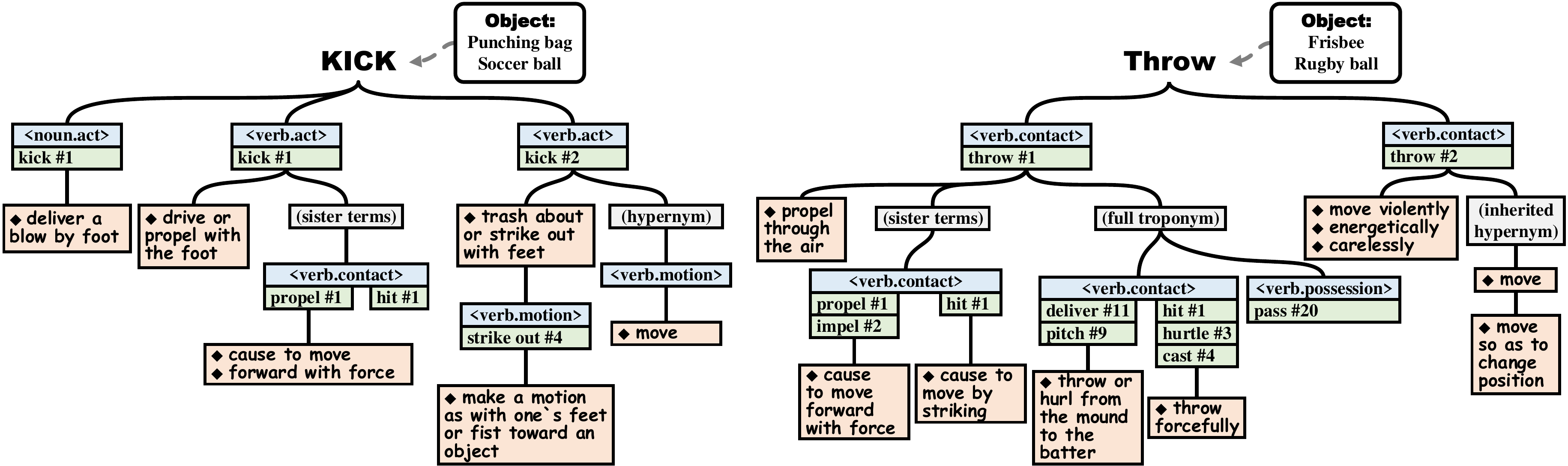}
  \caption{\textbf{Examples of utilizing WordNet \cite{miller1995wordnet} to explore potential actions which can be performed on specific objects}. WordNet groups words together based on specific senses. The interlinks between words in WordNet could be visually organized in the form of tree diagram. Verbs in WordNet are classified into $9$ major categories while nouns in $25$ categories. As for the word ``\textit{Kick}'' in WordNet, there are $6$ distinct senses when it is treated as a noun and $8$ different senses when it is a verb. In the tree diagram, ``$\langle noun.act \rangle$ kick \#1'' indicates that the first sense of noun ``kick'' belongs to ``act'' category. This semantic sense is glossed to ``deliver a blow by foot'' which could be an annotation phrase to affordance ``kick''. After filtering out irrelevant words semantic domains to affordance such as $\langle verb.emotion \rangle$, $\langle verb.competition \rangle$, \etc. we find that $\langle verb.act \rangle$, $\langle verb.motion \rangle$, $\langle verb.contact \rangle$ and $\langle noun.act \rangle$ are the most associated ones. Besides, we utilize the linguistic concepts in WordNet to explore richer expressions. In the tree diagram, ``\textit{hypernym}'' denotes more abstract and generic words, (\eg ``move'' is the hypernym of ``kick''), the expression ``\textit{sister terms}'' is used to represent a pair of synsets (sets of cognitive synonyms) which share a hypernym and ``\textit{troponym}'' indicates a ``manner'' relation between two verbs. With the leverage of these semantic relations, nodes in the tree diagrams could be employed as affordance annotations.}
  \label{FIG:anno}
\end{figure*}

Many approaches have been proposed at pixel level. In \cite{hu2016segmentation,li2018referring,margffoy2018dynamic}, the multimodal features from CNN and LSTM \cite{hochreiter1997long} are directly concatenated  together and input into the fully convolutional network to generate the final pixel-wise mask. These methods do not exploit the intra-modal and inter-modal relationships explicitly. More recent work uses self-attention and cross-attention mechanisms for linguistic and visual information. Ye \etal \cite{ye2021referring} proposed a cross-modal self-attention module to capture the long-range dependencies between linguistic and visual features. They can adaptively focus on essential words in the referring expression and crucial regions in the image. Hu \etal \cite{hu2020bi} designed a bi-directional relationship inferring network to model the relationship between linguistic and visual features. Liu \etal \cite{liu2021cross} proposed a model that first perceives all the entities in the image according to the entity and attribution words in the expression, then infers the location of the target object with the words that represent the relationship. Jing \etal \cite{jing2021locate} first gets the position prior of the referring object based on the language and image, then generates segmentation mask based on the position prior.  \par
Compared to the mentioned referring segmentation task, our proposed task has significant differences. At first, the inherent multiplicity feature of affordance leads to a much more significant variation in the visual representation of objects referred to by the same text than in traditional ones, which masks the alignment of linguistic and visual features more difficult. Secondly, our phrases only describe affordance without presenting entity words. Therefore, it is unable to utilize relationships between entities as in \cite{jing2021locate,liu2021cross,ye2021referring} to localize objects. We adopt short phrases to describe affordances rather than long sentences to meet practical scenarios. This disables us to leverage the text context information to capture the relationship between linguistic and visual features like the operations in \cite{liu2017recurrent,hu2020bi,ye2021referring,liu2021cross}.
The work mentioned above also utilized natural language to learn visual affordances \cite{mi2019object,mi2020intention}. Our work differs from them in the following ways.
Firstly, we consider the inherent multiplicity problem of affordance and involve richer indoor and outdoor scenes, which meet affordance's definition and are suitable for practical applications. Secondly, our model can generate a more precise pixel-level segmentation mask instead of a bounding box, limiting the ability to capture the inherent shape of objects. The accurate shape offers downstream tasks such as ``Robot Grasping'' richer geometric features to facilitate potential actions. 
In addition, technically, unlike their two-stage strategy, which relies heavily on the accuracy of intention extraction from natural language at first, we propose an end-to-end framework to enable multi-modal information to interact with each other adequately in a cyclic and bilateral manner. 

\section{Language Annotations} \label{anno}

The section shows the process to get our language annotations based on PAD dataset. The whole complete PAD dataset could be found at \url{https://github.com/lhc1224/OSAD_Net}. We describe details of the annotation process where we consider affordances from four perspectives with the assistance of WordNet, which is a hierarchical lexical database and could be explored at  \url{https://wordnet.princeton.edu/}. After that, we show some statistics of the proposed PAD-Language dataset. 

Instead of describing affordances with grammatically coherent sentences, we find it more effective to use several phrases closer to the actual application scenarios to describe the affordance. People tend to use short instructions to computers or robots in daily life rather than long sentences. Moreover, short phrases are more representative than complicated sentences to depict objects' affordances. \par 
As the inherent property of an object, the term ``\textit{affordance}'' is related to a set of possible actions that are able to manipulate objects. Therefore, the linguistic descriptions of affordance must be tightly relevant to these potential actions or the functionalities for human use. In addition, the environment in which objects are located and their appearance features may also affect the affordances of objects. Most of the affordances associated with our daily life are related to these aspects. Therefore, for better phrases descriptions, we consider text phrases from several different perspectives: \textbf{1)} The \textbf{actions} that can be potentially performed on the object. \textbf{2)} The \textbf{function} of the object. \textbf{3)} The \textbf{appearance features} related to the actions or functionalities. \textbf{4)} The \textbf{environment} that has capability to afford possible interactions between actors and bojects.

\textbf{\emph{\underline{1) Potential Actions:}}} Different from other properties of objects, affordance has one noteworthy distinction. That is, an object may have several different affordances, while different objects may have the same affordance. It is difficult to focus on this issue while describing affordance using natural language. To this end, we make the phrase descriptions based on the affordance categories rather than object categories in the PAD dataset. To explore more expressions for actions, we utilize a widely used  large lexical tool WordNet \cite{miller1995wordnet} to assist the annotation process. WordNet is a hierarchical lexical database that groups verbs, nouns, adjectives, and adverbs into sets of cognitive synonyms and \textit{synsets} associated by semantic relations. Other words or phrases expressing a similar sense could be found easily in WordNet for a specific word. To describe affordances, actions can be generally indicated by a ``\textit{synset}'' rather than individual verbs. Tree diagrams could represent the semantic relationship between words in WordNet. Two typical keyword centered tree diagram examples are shown in Fig. \ref{FIG:anno}. For a specific action, the phrases on nodes in the tree diagram can greatly enrich the descriptions for affordances. As shown in Fig. \ref{FIG:anno}, it is reasonable to consider that \textit{``soccer ball''} and \textit{``punching bag''} have the affordance \textit{``deliver a blow by foot''}, \textit{``drive or propel with foot''}, \textit{``strike out''}, \etc. and affordance \textit{Throw} could be extended to phrases such as \textit{propel through the air}, \textit{deliver} or \textit{pass}, \etc rather than a single word. It is worth noting that we adopt multiple verb tenses instead of only in original form to exhibit more diverse application scenarios and get more natural language phrases.

\textbf{\emph{\underline{2) Function:}}} Object function is an intrinsic property of an object independent of the users. Moreover, functionality understanding plays a vital role in human-machine interactions. Sense objects' function is essential to building a more intelligent computer vision system. Therefore, functions of an object are also included in our phrases annotations. For instance, the object ``\textit{Umbrella}'' has the function of ``\textit{sheltering from the wind and rain}'', object ``\textit{Knife}'' has the function of ``\textit{cut}'', object ``\textit{Drum}'' has the function of ``\textit{make sound}''. We annotate functions of objects in PAD dataset using simple phrases instead of involving in more details.

\textbf{\emph{\underline{3) Appearance Feature:}}} Visual appearance and geometric characteristics can be regarded as the physical basis of affordances. For instance, the middle of a \textit{cup} is \textit{depressed} resulting in its ability to ``\textit{hold water}''. ``\textit{soccer ball}'' is ``\textit{spherical}'' in appearance causing it to have the ability to ``\textit{roll}''. Besides, in practice, one may not know the specific category of an object but the affordances could be inferred by appearance features.

\textbf{\emph{\underline{4) Environment:}}} In the most accepted affordance definitions, the environment plays an important role. The term ``affordance'' is thought to reveal the complementary nature of the animal and the environment \cite{gibson1977theory}. In our textual annotations, we incorporate descriptions of the environment. Sometimes, specific affordances are available only when the object is located in a particular environment. For example, a ``\textit{soccer ball}'' generally only exhibits the affordance ``\textit{play}'' in an \textbf{outdoor} environment and ``\textit{chopsticks}'' are normally present at \textbf{the dining table} or \textbf{the kitchen}. After considering environment surrounding the object, the description of affordance becomes more complete. More examples could be found in Appendix. \par
Fig. \ref{FIG:stati} shows some overall statistics of the proposed PAD-L dataset which is constructed based on previous PAD dataset with containing $4,002$ images from 
$31$ affordance categories and $72$ object categories. We split these images into $75$\% training and $25$\% test. For a single image in the PAD dataset, we randomly select a set of phrases (the number is $4$ in this paper) from the candidate annotations to build a new extended version with text information considered. The statistic shows that PAD-L contains rich phrases in a variety of scenarios. 

\begin{figure*}[!t]
  \begin{minipage}[b]{0.28\textwidth} 
    \centering
    \small
    \renewcommand{\arraystretch}{1.}
    \renewcommand{\tabcolsep}{4.pt}
        \begin{tabular}{l|c}
    \hline
    \Xhline{2.\arrayrulewidth}
        \rowcolor{mygray}
        \textbf{Statistics} &  \textbf{Overall}  \\
    \hline
    \Xhline{2.\arrayrulewidth}
        \# affordances &  31   \\
    \hline
        \# objects  &    72  \\
    \hline
        \# images  &     4,002      \\
        \multicolumn{1}{l|}{\#\# Train Set} & 3,202\\
        \multicolumn{1}{l|}{\#\# Test Set}  & 800 \\
    \hline
        \# phrases &    1,447    \\
    \hline
        \# words &     959    \\
    \hline
        \# words per phrases &  2.36 \\
    \hline
        \# phrases per affordance  & 46.68 \\
    \hline
    \Xhline{2.\arrayrulewidth}
    \end{tabular}
  \end{minipage}
  \begin{minipage}[htb]{0.46\textwidth}
  \centering
  \includegraphics[width=0.99\textwidth]{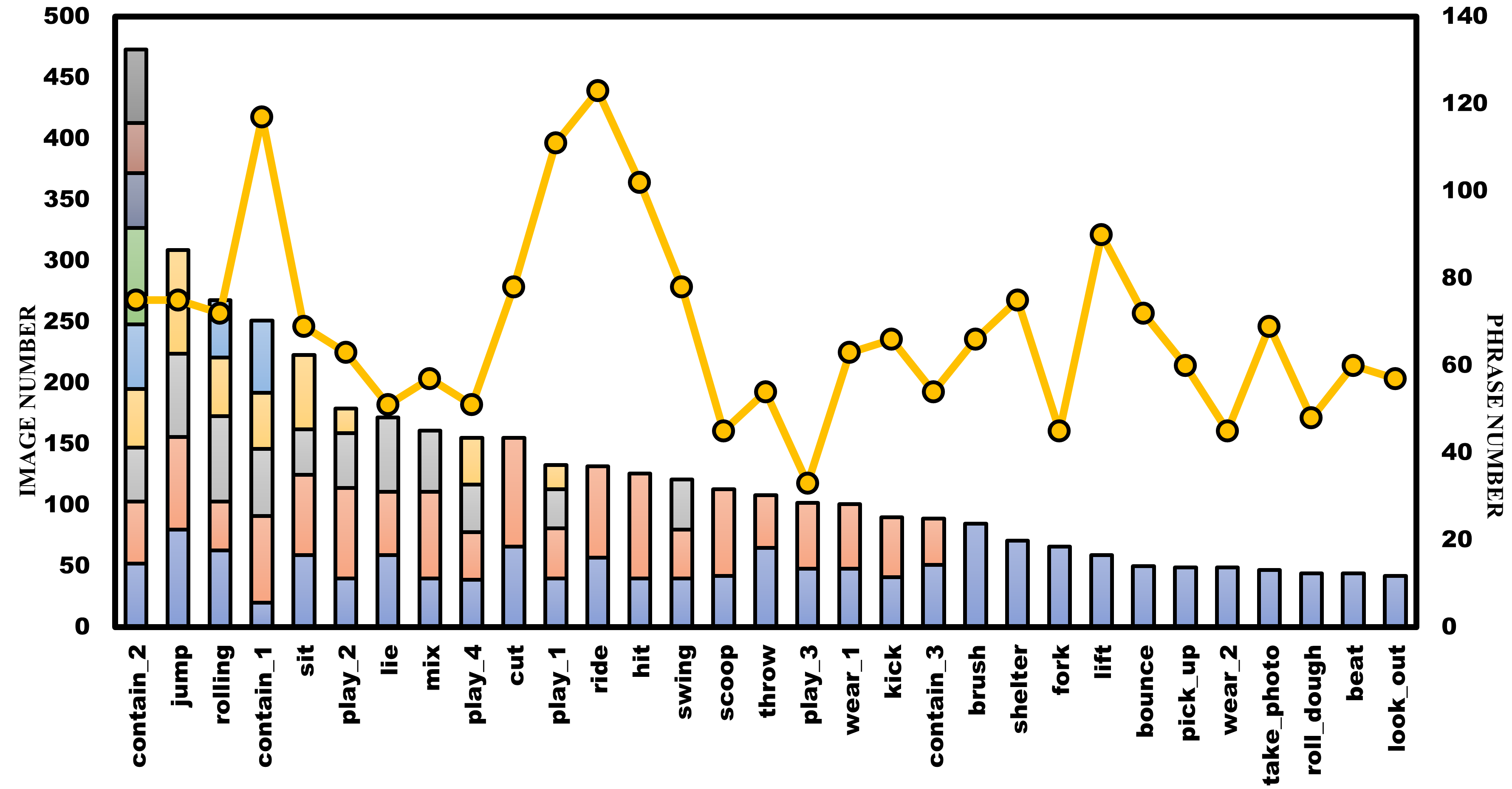}
  \end{minipage}
  \begin{minipage}[htb]{0.25\textwidth}
  \centering
    \includegraphics[width=0.99\textwidth]{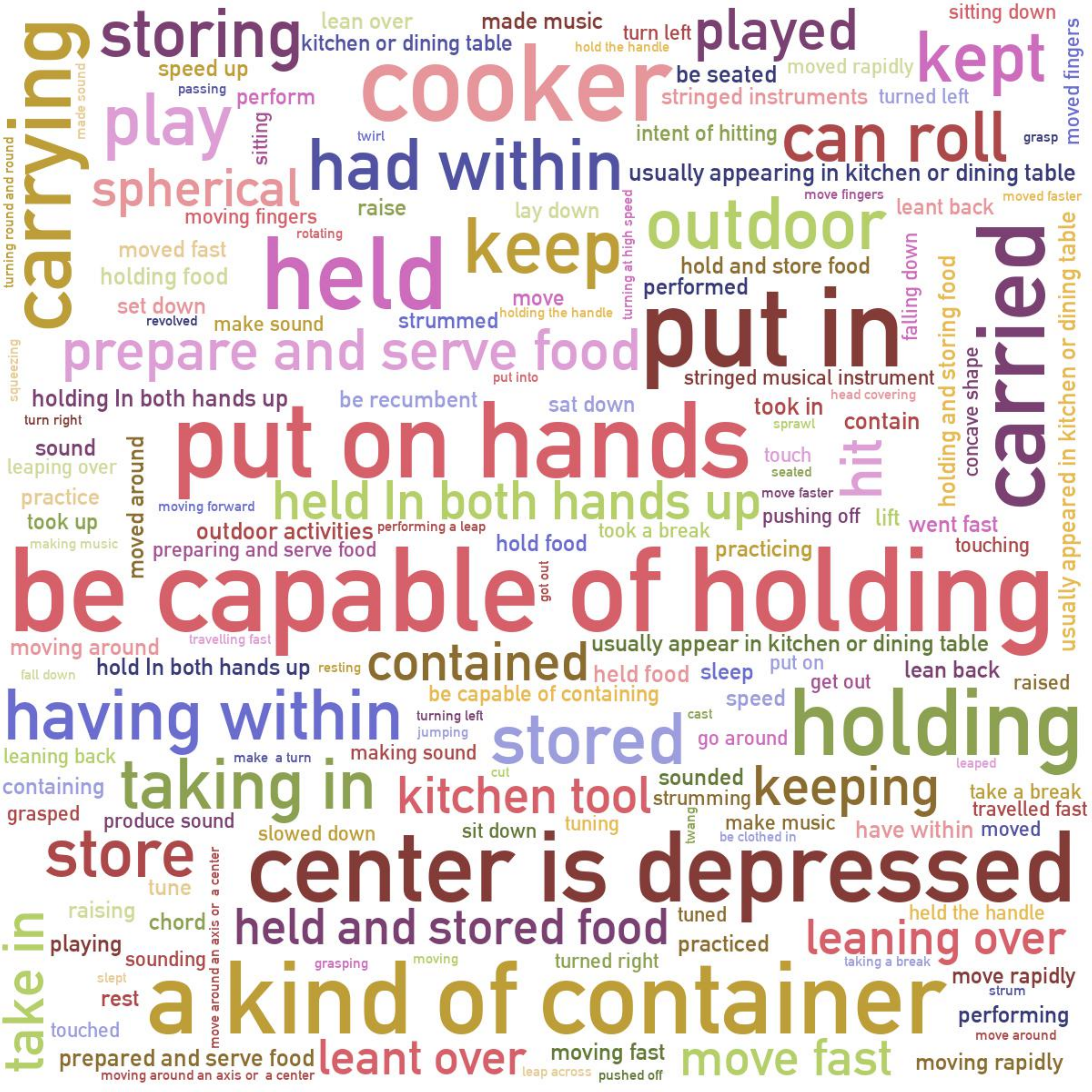}
  \end{minipage}
  \caption{\textbf{Statistics of PAD-L.} The table on the left shows the statistics of all data. The middle chart illustrates the image number and phrase number according to affordance categories. The bars represent image numbers, with every part in the bar indicating the image number of every object category, and the line chart shows phrase numbers. The cloud of phrases on the right has the font sizes proportional to the square root of frequencies in the proposed PAD-L.}
  \label{FIG:stati}
\end{figure*}


\section{Method} \label{method}

\subsection{Problem Description}

Taking a set of affordance descriptions $P$=$\{P_1, P_2,..., P_n\}$ that describes affordance $A_m$ and an affordance related image $I$ which contains multiple objects $\{O_1, O_2, ..., O_n\}$, the phrase-based affordance detection task aims to get the segmentation mask $M$ of the object $O_m$ which can afford the affordance $A_m$ in the image. We define the input as $n$ query phrases and an image $I$ in each batch.\par
We need a vision encoder and language encoder to encode visual and linguistic features to align and fuse information from two distinct modalities. Then, the result features are input to a module to learn the consistency between the two modalities. After adequate alignment and fusion,
we use a segmentation module to generate the final masks.

\subsection{Visual and Linguistic Features Extraction}
\label{Overall}

As shown in Fig. \ref{FIG:model} (a), our model takes a set of phrases and an affordance related image as inputs. In the image branch, a CNN backbone (\eg ResNet101\cite{7780459}) is used to extract multi-level image features. The output of the 3rd, 4th and 5th stages of CNN backbone are denoted as $\{I_3, I_4, I_5\}$ with channel dimension of $512$, $1024$ and $2048$, respectively. Afterwards, $1 \times 1$ convolution layers are employed to transform the multi-level visual features to the same size of  $\mathbb{R}^{H\times W \times C_{I}}$.\par
In the textual branch, the language features $L=\{L_1, L_2,..., L_n\}$ is extracted using a language encoder (\eg LSTM\cite{6795963}), where  $n$ is the number of phrases. The parameters of embedding layer is initialized using GloVe word embeddings \cite{pennington-etal-2014-glove}. After encoding by the language encoder, the resulting linguistic features are applied in a max pooling operation and get a global language representation $L_0 \in \mathbb{R}^{C_l}$ where $C_l$ is the dimension of the language feature to interacts with the multi-level visual features and feed into the proposed \textit{Cyclic Interaction Module}.\par
Afterwards, a bilinear fusion \cite{ben2017mutan} is adopted to fuse different level visual features with linguistic feature $L_0$. Following the prior work in referring segmentation\cite{Liu2017RecurrentMI}\cite{ye2021referring}\cite{hu2016segmentation}, to incorporate more spatial information, we concatenate a 8-D spatial coordinate feature which is denoted as $P\in \mathbb{R}^{H \times W \times 8}$ with the resulting fused multi-model feature before to get final fused features $\{F^3_0, F^4_0, F^5_0 \} \in \mathbb{R}^{H \times W \times (C_I + C_l + 8)}$, which could be defined as following:
\begin{equation}
\{ F^i_0 = concat(f(I_i, L_0), P)\}_{i=3,4,5}, \label{eq:1}
\end{equation}
where $concat(\cdot , \cdot)$ represents the concatenation operation along the channel dimension and $f$ denotes bilinear fusion operation. In this paper, $H=40$, $W=40$.

\subsection{Cyclic Interaction Module}
\label{CIM}

Compared to traditional vision-language tasks, the apparent difference of objects referred to by the same language descriptions in different images could be significant because of the multiplicity property of affordance. \par
The vast differences in visual appearance lead to substantial divergence in the distribution of textual and visual features in feature space. To generate accurate and consistent representations of the target object and the given affordance description phrases, the feature representation for one modality is enhanced several times adaptively guided by the other modality in \textit{Cyclic Interaction Module (CIM)}, which is indicated in Fig. \ref{FIG:model} (a). CIM consists of bilateral interaction operations between the two modalities to learn the consistency step by step, which leads to adequate fusion and alignment.\par 
Specifically, we propose a \textit{Vision guide Language Module} (VLM) to enhance the linguistic feature representation with the guidance of visual features and a \textit{Language guide Vision Module} (LVM)) to get improved visual feature representations. The cyclic interaction process is illustrated as following, ($i=3,4,5$ and $j,k \in \{3,4,5\} \backslash \{i\}$ in the equations):
\begin{align}
L^i_1 & = VLM(L_0, F^i_0),\label{eq:2} \\ 
F^i_1 & = LVM(L^i_1, F^i_0, F^j_0, F^k_0), \label{eq:3} \\
L^i_2 & = VLM(L^i_1, F^i_1),\label{eq:4} \\
F^i_2 & = LVM(L^i_2, F^i_1, F^j_1, F^k_1). \label{eq:5}
\end{align}

\subsection{Vision Guide Language Module}
\label{VLM}

\begin{figure*}[t]
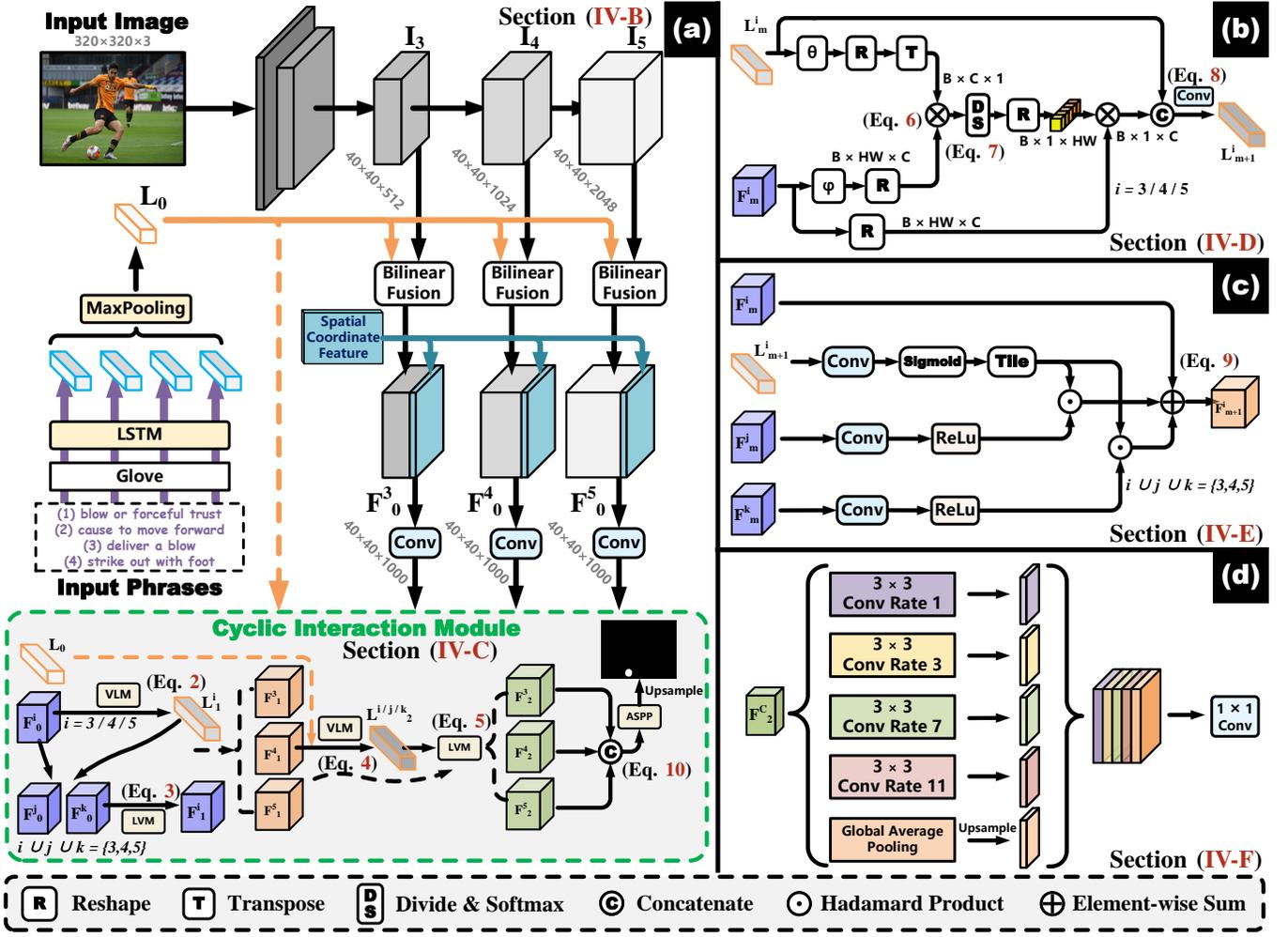

	\centering
		\begin{overpic}[width=0.99\linewidth]{figure/model}
		\put(11.5,19.6){\footnotesize{\textbf{(Eq.~\ref{eq:2})}}}
		\put(9.5,11){\footnotesize{\textbf{(Eq.~\ref{eq:3})}}}
		\put(25,13.2){\footnotesize{\textbf{(Eq.~\ref{eq:4})}}}
		\put(34,16.5){\footnotesize{\textbf{(Eq.~\ref{eq:5})}}}
		\put(68,64){\footnotesize{\textbf{(Eq.~\ref{eq:6})}}}
		\put(74.5,61.5){\footnotesize{\textbf{(Eq.~\ref{eq:7})}}}
		\put(92,67.5){\footnotesize{\textbf{(Eq.~\ref{eq:8})}}}
		\put(93,45){\footnotesize{\textbf{(Eq.~\ref{eq:9})}}}
		\put(49,12.7){\footnotesize{\textbf{(Eq.~\ref{eq:10})}}}
		
		\put(39.3,72){\normalsize{\textbf{Section~(\ref{Overall})}}}
		\put(27,22){\normalsize{\textbf{Section~(\ref{CIM})}}}
		\put(87.5,54.2){\normalsize{\textbf{Section~(\ref{VLM})}}}
		\put(87.5,31.2){\normalsize{\textbf{Section~(\ref{LVM})}}}
		\put(87.5,5.5){\normalsize{\textbf{Section~(\ref{SegModule})}}}
		\end{overpic}
		\caption{\textbf{The architecture of our proposed CBCE-Net.} CBCE-Net first uses DeepLab Resnet101 \cite{chen2017deeplab} and LSTM \cite{6795963} to extract multi-level visual and linguistic features, respectively. Subsequently, combining spatial coordinate, multi-level multi-modal features are generated through bilinear fusion operations (see Section \ref{Overall} for details). Afterwards, fused features are fed into CIM module to enhance the semantic consistency in a cyclic and bilateral manner (See Section \ref{CIM} for details). In CIM module, we design a VLM module (see Section \ref{VLM}) and a LVM module (see Section \ref{LVM}) to update visual and linguistic features bilaterally with the guidance of each other. VLM is showed in part (b) in the top right corner and LVM is illustrated in part (c). Note that in LVM, the original feature is showed at the top left corner which denoted as $F^i_m$ and the updated feature is at the output denoting as $F^i_{m+1}$. At last, a ASPP module (shown in part (d)) receives the final concatenated fused features and generate predicted masks (see Section \ref{SegModule}).
		}
	\label{FIG:model}
\end{figure*}

The architecture of the proposed VLM is illustrated in Fig. \ref{FIG:model} (b). To update the language feature representation $L^i_{m+1}$, we leverage the previous fused feature $F^i_m$ to guide the transformation of the previous language feature $L^i_m$ in VLM. \par
For a language feature $L^i_m \in \mathbb{R}^{C_l}$ and fused visual feature $F^i_m \in \mathbb{R}^{H \times W \times C_v}$, we can compute the element-wise correlations using inner product:
\begin{equation}
    S^i_m = \phi(F^i_m)\theta(L^i_m)^T, \label{eq:6}
\end{equation}
where $\theta$ and $\phi$ are $1 \times 1$ convolution layers to transform the feature to have the same dimensions where $\theta(L^i_m) \in \mathbb{R}^{1 \times C}$, $\phi(F^i_m) \in \mathbb{R}^{HW \times C}$.
The affinity map $S^i_m \in \mathbb{R}^{HW \times 1}$ captures the correlation information of the given features. Then we employ \textit{scale} and \textit{softmax} operations following the scaled dot-product attention practice in \cite{vaswani2017attention} to normalize and reshape the affinity map to produce global affinity attention map $A^i_m \in \mathbb{R}^{1 \times HW}$. This process is shown as following:
\begin{equation}
    A^i_m = \text{Softmax}(\frac{S^i_m}{\sqrt{C}}). \label{eq:7}
\end{equation}
Afterwards, we multiply the reshaped original fused visual feature $F^i_m$ to the resulting attention map to generate the attention feature map $A^i_c \in \mathbb{R}^{1 \times C}$ along the channel dimension. Finally, to get the final updated language feature representation $L^i_{m+1} \in \mathbb{R}^{1 \times 1 \times C}$, we concatenate the original language feature $L^i_m$ to $A^i_c$ followed by a convolution layer and a $L_2$ normalization operation as shown following:
\begin{equation} \label{eq:8}
    L^i_{m+1} = ||conv(concat(L^i_m, A^i_c))||_2,
\end{equation}
where $conv$, $concat(\cdot)$, $||\cdot||_2$ denotes $1 \times 1$ convolution, $concatenate$ and $L_2$ normalization operations, respectively. 

\subsection{Language Guide Vision Module}
\label{LVM}

Previous work \cite{jing2021locate,li2018referring,yu2017multi} on vision-language tasks demonstrates that the information exchange among multi-level features benefits the vision and language interaction process a lot. Therefore, leveraging multi-level information, we propose a novel \textit{Language Guide Vision Module} to update the visual features under the guidance of linguistic features. The operation is shown in Fig. \ref{FIG:model} (c). 

The updated linguistic feature $L^i_{m+1} \in \mathbb{R}^{1 \times 1 \times C}$ contains rich multimodal context information of  $F^i_m$. We utilize $L^i_{m+1}$ to select the relevant information from other two level features $F^j_m$ and $F^k_m$ after necessary transformations. The final aggregated global context feature $F^i_{m+1}$ is obtained by adding $F^i_m$ and relevant information from other two levels:
\begin{equation}\label{eq:9}
    F^i_{m+1} = F^i_m + \sum_{k\in \{3,4,5\} \backslash i}  \sigma(conv(L^i_{m+1})) \odot F^k_m, 
\end{equation}
where $\sigma(\cdot)$ denotes \textit{sigmoid} function.

\subsection{Segmentation Module}
\label{SegModule}

The segmentation module aims to produce the final fine segmentation mask. At first, we obtain a concatenated feature $F^C_2$ which contain multi-level information:
\begin{equation} \label{eq:10}
    F^C_2 = concat(F^3_2, F^4_2, F^5_2).
\end{equation}
Next, as shown in Fig. \ref{FIG:model} (d), we utilize a ASPP module \cite{chen2017deeplab} to capture multi-scale information. The structure of ASPP consists of five parallel sub-networks. The first one learns global information by employing \textit{global average pooling} operation while the remaining four branches apply atrous convolutions with multiple dilation rates of $\{1, 3, 7, 11\}$ respectively. 
In the parallel branches, the depthwise separable convolution is applied to reduce the model complexity. After that, the multi scale features are concatenated together. Finally, a $1\times 1$ convolution and a \textit{upsample} operation to are adopted to generate the final fine mask $P_m$ with the same resolution and channel dimension as the input image. \par
During training, we adopt the Sigmoid Binary Cross Entropy (BCE) loss as a minimized objective, which is defined on the predicted output $P_{mask}$ and the ground truth segmentation mask $G$ as following:
\begin{equation}
L = \sum_{i=1}^{H \times W} G(i)log(P_{mask}(i)) + (1-P_{mask}(i))log(1-G(i)), \label{eq:11}
\end{equation}
where $i$ is the elements of the ground-truth mask and $H \times W$ denotes the size of the ground-truth mask.

\begin{table*}[t]
\caption{\textbf{The experimental results of $10$ models} (\textcolor[rgb]{0.8,0.0,0.3}{\textbf{BASNet}}\cite{Qin_2019_CVPR} , \textcolor[rgb]{0.8,0.0,0.3}{\textbf{CPD}}\cite{Wu_2019_CVPR}, \textcolor[rgb]{0.99,0.5,0.0}{\textbf{OSAD}}\cite{Oneluo}, \textcolor[rgb]{0.99,0.5,0.0}{\textbf{OAFFD}}\cite{zhao2020object},
\textcolor[rgb]{0.4,0.0,0.99}{\textbf{PSPNet}}\cite{zhao2017pspnet}, \textcolor[rgb]{0.4,0.0,0.99}{\textbf{DeepLabv3+ (DLabV3+)}}\cite{chen2017rethinking}
\textcolor[rgb]{0.0,0.8,0.0}{\textbf{CMSA}}\cite{ye2021referring}, \textcolor[rgb]{0.1,0.8,0.1}{\textbf{BRINet}}\cite{hu2020bi}, \textcolor[rgb]{0.1,0.8,0.1}{\textbf{CMPC}}\cite{liu2021cross}) in terms of five metrics (IoU~$(\uparrow)$, $F_{\beta}$~$(\uparrow)$, $E_\phi$~$(\uparrow)$, CC~$(\uparrow)$, and MAE~$(\downarrow)$). \textbf{Bold} and \underline{underline} indicate the best and the second-best scores, respectively.}
  \label{Table:2}
\centering
  \renewcommand{\arraystretch}{1.}
  \renewcommand{\tabcolsep}{3.pt}
  \begin{tabular}{c||cc|cc|cc|ccc||c}
    \hline
    \Xhline{2.\arrayrulewidth}
    \rowcolor{mygray}
    \textbf{Methods}  & \textcolor[rgb]{0.8,0.0,0.3}{\textbf{BASNet}} \textbf{\cite{Qin_2019_CVPR}} & \textcolor[rgb]{0.8,0.0,0.3}{\textbf{CPD}} \textbf{\cite{Wu_2019_CVPR}} & \textcolor[rgb]{0.99,0.5,0.0}{\textbf{OSAD}} \textbf{\cite{Oneluo}} & \textcolor[rgb]{0.99,0.5,0.0}{\textbf{OAFFD}} \textbf{\cite{zhao2020object}} & \textcolor[rgb]{0.4,0.0,0.99}{\textbf{PSPNet}} \textbf{\cite{zhao2017pspnet}} & \textcolor[rgb]{0.4,0.0,0.99}{\textbf{DLabV3+}} \textbf{\cite{chen2017rethinking}} & \textcolor[rgb]{0.0,0.8,0.0}{\textbf{CMSA}} \textbf{\cite{ye2021referring}} & \textcolor[rgb]{0.1,0.8,0.1}{\textbf{BRINet}} \textbf{\cite{hu2020bi}} & \textcolor[rgb]{0.1,0.8,0.1}{\textbf{CMPC}} \textbf{\cite{liu2021cross}}     & \textbf{Ours} \\
    \hline
    \rowcolor{mygray}
    Year & 2019  & 2019 & 2021 & 2020 & 2017 & 2018 & 2021 & 2020  & 2021  & $\backslash$ \\
    \hline
    \Xhline{2.\arrayrulewidth}
     \textbf{IoU ($\uparrow$)}  & $0.491$  & $0.496$  & $0.554$ & $0.439$ & $0.464$ & $0.509$ & $0.571$ & \underline{$0.579$} & \underline{$0.579$}   & \bm{$0.593$}  \\
     \textbf{$E_{\phi}$ ($\uparrow$)} & $0.752$ & $0.744$ & $0.777$ & $0.714$ & $0.692$ & $0.761$ & $0.799$ & $0.793$ & \underline{$0.806$}  & \bm{$0.822$}   \\
     \textbf{CC ($\uparrow$)} & $0.557$ & $0.626$ & $0.662$ & $0.565$ & $0.573$ & $0.638$ & \underline{$0.711$} & $0.710$ & $0.706$  & \bm{$0.713$} \\
     \textbf{MAE ($\downarrow$)}   & $0.086$ & $0.083$ & $0.083$ & $0.098$ & $0.138$ & $0.064$ & $0.063$ & \underline{$0.061$} & $0.062$   & \bm{$0.061$}     \\
     \textbf{$F_{\beta}$ ($\uparrow$)}  & $0.571$ & $0.573$ & $0.630$ & $0.521$ & $0.503$ & $0.631$ & $0.644$ & \underline{$0.653$}  & $0.650$  & \bm{$0.665$} \\
    \hline
    \Xhline{2.\arrayrulewidth}
    \end{tabular}
    \label{objective results}
  \end{table*}

\begin{table*}[t]
    \centering
    \renewcommand{\arraystretch}{1.}
    \renewcommand{\tabcolsep}{9.pt}
    \caption{\textbf{The results of different methods on the PAD-L for each affordance category.} IoU is used as the evaluation metric. \textbf{Bold} and \underline{underline} indicate the best and the second-best scores, respectively.}
  \begin{tabular}{c||cc|cc|cc|ccc||c}
    \hline
    \Xhline{2.\arrayrulewidth}
\textbf{Classes}  & \makecell[c]{\textcolor[rgb]{0.8,0.0,0.3}{\textbf{BASNet}} \\ \textbf{\cite{Qin_2019_CVPR}}} & \makecell[c]{\textcolor[rgb]{0.8,0.0,0.3}{\textbf{CPD}} \\ \textbf{\cite{Wu_2019_CVPR}}} & \makecell[c]{\textcolor[rgb]{0.99,0.5,0.0}{\textbf{OSAD}} \\ \textbf{\cite{Oneluo}}} & \makecell[c]{\textcolor[rgb]{0.99,0.5,0.0}{\textbf{OAFFD}} \\ \textbf{\cite{zhao2020object}}} & \makecell[c]{\textcolor[rgb]{0.4,0.0,0.99}{\textbf{PSPNet}} \\ \textbf{\cite{zhao2017pspnet}}} & \makecell[c]{\textcolor[rgb]{0.4,0.0,0.99}{\textbf{DLabV3+}} \\ \textbf{\cite{chen2017rethinking}}} & \makecell[c]{\textcolor[rgb]{0.0,0.8,0.0}{\textbf{CMSA}} \\ \textbf{\cite{ye2021referring}}} & \makecell[c]{\textcolor[rgb]{0.1,0.8,0.1}{\textbf{BRINet}} \\ \textbf{\cite{hu2020bi}}} & \makecell[c]{\textcolor[rgb]{0.1,0.8,0.1}{\textbf{CMPC}} \\ \textbf{\cite{liu2021cross}}}     & \textbf{Ours} \\
   \hline
    \Xhline{2.\arrayrulewidth}
\rowcolor{mygrayd}
\textbf{Beat} & $0.548$ & $0.625$ & $0.808$ & $0.562$ & $0.572$ & $0.671$  & \underline{$0.813$} & \bm{$0.835$}  & $0.779$  & $0.766$    \\

\textbf{Bounce} & $0.362$ & $0.524$ & $0.601$ & $0.376$ & $0.427$ & $0.564$ & $0.606$  & $0.642$ & \bm{$0.652$} & \underline{$0.616$} \\
\rowcolor{mygrayd}
\textbf{Brush} & $0.275$ & $0.369$ & $0.427$ &  $0.267$ & $0.292$ & $0.395$ & \underline{$0.449$}  & \bm{$0.450$} & $0.440$ & $0.443$   \\ 

\textbf{Contain-1} & $0.290$ & $0.393$ & $0.463$ & $0.313$ & $0.355$ & $0.404$ & \underline{$0.493$} & $0.489$  & $0.481$ & \bm{$0.508$ }      \\
\rowcolor{mygrayd}
\textbf{Contain-2} & $0.447$ & $0.518$ & $0.573$ & $0.449$ & $0.483$ & $0.539$ & $0.608$ & $0.609$  & \underline{$0.618$} & \bm{$0.634$} \\

\textbf{Contain-3} & $0.485$ & $0.543$ & $0.593$ & $0.482$ & $0.511$ & $0.555$ & $0.631$  & $0.629$ & \underline{$0.635$} & \bm{$0.656$}    \\
\rowcolor{mygrayd}
\textbf{Cut} & $0.448$ & $0.511$ & $0.557$ & $0.446$ & $0.471$ & $0.524$ & $0.594$ & $0.587$ & \underline{$0.595$} & \bm{$0.621$}          \\

\textbf{Fork} & $0.433$ & $0.490$ & $0.538$ &  $0.431$ & $0.455$ & $0.507$ & \underline{$0.575$} & $0.567$ & $0.574$ & \bm{$0.603$}         \\
\rowcolor{mygrayd}
\textbf{Hit} & $0.420$ & $0.475$  & $0.531$ & $0.421$ & $0.446$ & $0.500$ & $0.561$ & $0.552$ & \underline{$0.562$} & \bm{$0.590$}      \\

\textbf{Jump} & $0.395$ & $0.438$ & $0.502$ &  $0.388$ & $0.404$ & $0.458$  & $0.523$ & $0.520$ & \underline{$0.526$} & \bm{$0.556$}     \\
\rowcolor{mygrayd}
\textbf{Kick}  &  $0.409$ & $0.450$ & $0.516$ & $0.400$ & $0.410$ & $0.471$ & $0.531$    & $0.533$  & \underline{$0.536$} & \bm{$0.567$} \\

\textbf{Lie} & $0.442$ & $0.476$ & $0.541$ & $0.425$  & $0.439$ & $0.491$ & $0.547$ & $0.553$ & \underline{$0.554$} & \bm{$0.579$} \\
\rowcolor{mygrayd}
\textbf{Lift} & $0.445$ & $0.480$ & $0.546$ &  $0.429$ & $0.443$ & $0.494$  & $0.549$ & $0.554$ & \underline{$0.557$} & \bm{$0.581$}            \\

\textbf{Look Out} & $0.448$ & $0.484$ & $0.549$ & $0.433$ & $0.447$ & $0.499$ & $0.552$ & $0.557$ & \underline{$0.558$} & \bm{$0.583$}    \\
\rowcolor{mygrayd}
\textbf{Mix} & $0.465$ & $0.488$ & $0.542$ & $0.428$ & $0.449$ & $0.488$ & $0.541$ & $0.541$ & \underline{$0.547$} & \bm{$0.568$}        \\
\textbf{Pick Up} & $0.469$ & $0.488$ & $0.541$ &  $0.427$ & $0.448$ & $0.486$ & $0.538$ & $0.538$ & \underline{$0.546$} & \bm{$0.567$}  \\
\rowcolor{mygrayd}
\textbf{Play-1} & $0.483$ & $0.498$ & $0.553$ & $0.435$ & $0.461$  & $0.503$ & $0.551$ & $0.551$ & \underline{$0.559$}  & \bm{$0.578$}  \\

\textbf{Play-2}  & $0.497$ & $0.513$ & $0.563$ & $0.447$ & $ 0.476$  & $0.517$ & $0.561$ & $0.564$ & \underline{$0.570$} & \bm{$0.589$}\\
\rowcolor{mygrayd}
\textbf{Play-3}  & $0.493$ & $0.519$ & $0.572$ & $0.452$ & $0.483$ & $0.525$ & $0.570$ & $0.572$ & \underline{$0.578$} & \bm{$0.596$}    \\

\textbf{Play-4} & $0.499$ & $0.519$ & $0.574$ & $0.451$  & $0.484$ & $0.526$ & $0.571$ & \underline{$0.579$} & $0.578$ & \bm{$0.596$} \\
\rowcolor{mygrayd}
\textbf{Ride} & $0.502$ & $0.518$ & $0.575$ & $0.455$  & $0.486$ & $0.528$  & $0.574$  & $0.575$ & \underline{$0.580$}   & \bm{$0.597$} \\

\textbf{Roll Dough}  & $0.500$ & $0.518$ & $0.576$ & $0.454$  & $0.486$ & $0.530$ & $0.574$ & $0.576$ & \underline{$0.580$} & \bm{$0.598$} \\
\rowcolor{mygrayd}
\textbf{Rolling} & $0.500$ & $0.515$ & $0.570$ &  $0.456$ & $0.479$ & $0.525$ & $0.579$ & $0.580$ & \underline{$0.588$} & \bm{$0.603$}     \\

\textbf{Scoop}  & $0.501$ & $0.511$ & $0.564$ & $0.451$ & $0.473$ & $0.517$ & $0.572$ & $0.573$ & \underline{$0.580$} & \bm{$0.596$}     \\
\rowcolor{mygrayd}
\textbf{Shelter} & $0.495$ & $0.504$ & $0.556$ &  $0.445$ & $0.465$ & $0.514$ & $0.574$ & $0.575$ & \underline{$0.581$}  & \bm{$0.595$}  \\

\textbf{Sit} & $0.499$ & $0.505$ & $0.559$ &  $0.446$ & $0.469$ & $0.516$  & $0.572$ & $0.574$ & \underline{$0.581$} & \bm{$0.595$} \\
\rowcolor{mygrayd}
\textbf{Swing} & $0.494$ & $0.499$ & $0.555$  & $0.440$ & $0.461$  & $0.507$ & $0.572$ & $0.574$ & \underline{$0.581$} & \bm{$0.596$}    \\

\textbf{Take Photo} & $0.494$ & $0.499$ & $0.555$ &  $0.441$ & $0.461$ & $0.508$ & $0.573$ & $0.574$ & \underline{$0.581$} & \bm{$0.596$} \\
\rowcolor{mygrayd}
\textbf{Throw} & $0.491$ & $0.498$ & $0.555$ & $0.438$ & $0.458$ & $0.510$ & $0.571$ & $0.576$ & \underline{$0.580$} & \bm{$0.594$} \\

\textbf{Wear-1} & $0.492$ & $0.499$ & $0.557$ & $0.441$  & $0.462$  & $0.513$ & $0.574$ & $ 0.581$ & \underline{$0.584$} & \bm{$0.597$}   \\
\rowcolor{mygrayd}
\textbf{Wear-2} & $0.491$ & $0.496$ & $0.553$ & $0.439$ & $0.459$ & $0.510$ & $0.571$ & \underline{$0.579$} & $0.577$ & \bm{$0.593$}    \\
    \hline
    \Xhline{2.\arrayrulewidth}
    \end{tabular}
    \label{Table:IoU}
\end{table*}

\section{Experiments} \label{exp}

\begin{figure*}[t]
	\centering
	\small
		\begin{overpic}[width=0.99\linewidth]{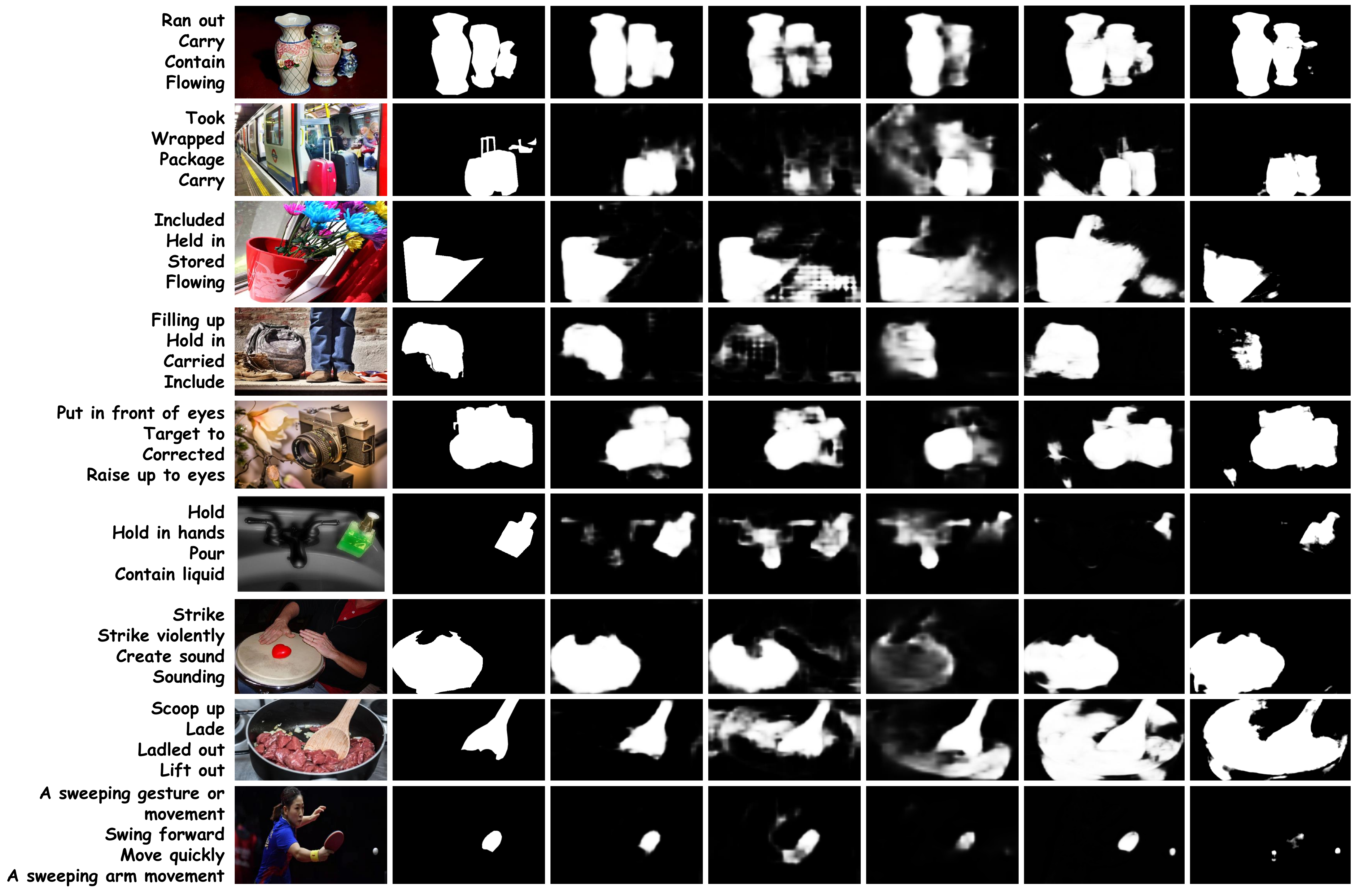}
		\put(6.5,-1){\textbf{Phrases}}
		\put(20,-1){\textbf{Image}}
		\put(33,-1){\textbf{GT}}
		\put(44,-1){\textbf{Ours}}
		\put(53.5,-1){\textcolor[rgb]{0.1,0.8,0.1}{\textbf{CMPC}} \textbf{\cite{liu2021cross}}}
		\put(66,-1){\textcolor[rgb]{0.8,0.0,0.3}{\textbf{CPD}} \textbf{\cite{Wu_2019_CVPR}}}
		\put(77,-1){\textcolor[rgb]{0.99,0.5,0.0}{\textbf{OSAD}} \textbf{\cite{Oneluo}}}
		\put(87,-1){\textcolor[rgb]{0.4,0.0,0.99}{\textbf{DeepLabV3+}} \textbf{\cite{chen2017rethinking}}}
		\end{overpic}
		\caption{\textbf{Visual results obtained by different models}, including \textcolor[rgb]{0.1,0.8,0.1}{\textbf{CMPC}} \cite{liu2021cross}, \textcolor[rgb]{0.8,0.0,0.3}{\textbf{CPD}} \cite{Wu_2019_CVPR}, \textcolor[rgb]{0.99,0.5,0.0}{\textbf{OSAD}} \cite{Oneluo} and \textcolor[rgb]{0.4,0.0,0.99}{\textbf{DeepLabV3+}} \cite{chen2017rethinking}.}
	\label{FIG:visual}
\end{figure*}
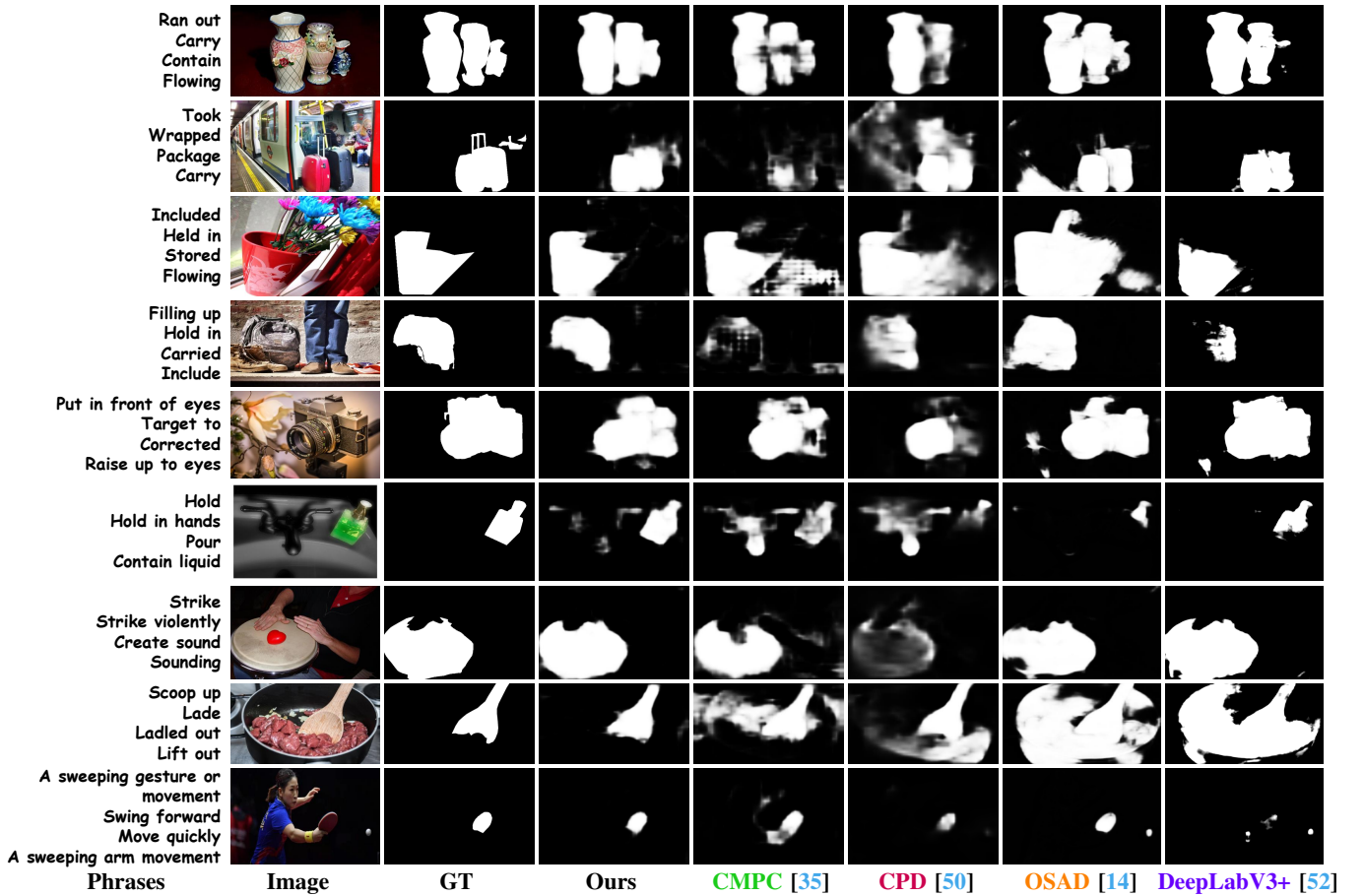

This section elaborates on the experiments' details, including experiment settings, results, and analysis. Section \ref{settings} presents the evaluation metrics and comparison methods we choose. 
In Section \ref{implement}, we describe the implementation details of our experiments. Section \ref{analysis} analysis the results of our model. Section \ref{ablation} demonstrates the ablation study.

\subsection{Settings} \label{settings}

We choose five broadly used metrics to comprehensively evaluate the performance of different methods, \ie, Intersection over Union (IoU), F-measure ($F_{\beta}$), E-measure ($E_{\phi}$), Pearson's Correlation Coefficient (CC), and Mean Absolute Error (MAE). More details could be found in the Appendix.\par
To illustrate the superiority of our model, we compare several different kinds of methods, which involve \textbf{two} \textcolor[rgb]{0.8,0.0,0.3}{\textbf{Salient Detection} models (BASNet, CPD)}, \textbf{two} \textcolor[rgb]{0.99,0.5,0.0}{\textbf{Affordance Detection} models (OSAD-Net, OAFFD)}, \textbf{two} \textcolor[rgb]{0.4,0.0,0.99}{\textbf{Semantic Segmentation} models (PSPNet, DeepLabV3+)}, and \textbf{three} \textcolor[rgb]{0.1,0.8,0.1}{\textbf{Referring Segmentation} models (CMSA, BRINet, CMPC)}. More details could be found in the Appendix.\par 

\subsection{Implementation Details} \label{implement}

Our method is implemented using TensorFlow. For visual feature extraction, we choose DeepLab-ResNet101 network \cite{chen2017deeplab} which is pre-trained on PASCAL-VOC dataset \cite{everingham2010pascal} as the backbone\footnote{The pretrained DeepLab-ResNet101 model can be dowloaded at \href{https://drive.google.com/drive/folders/0B_rootXHuswsZ0E4Mjh1ZU5xZVU?resourcekey=0-9Ui2e1br1d6jymsI6UdGUQ}{Link}.}. We use the outputs of the DeepLab blocks \textit{Res3}, \textit{Res4} and \textit{Res5} as the input multi-level visual features $\{I_3, I_4, I_5\}$. The parameters of the backbone keep are fixed in the training phase. During training, the input images are randomly clipped from $360 \times 360$ to $320 \times 320$ with a random horizontal flipping. The multi-level visual feature dimension $C_I$ is set to be 1000 in this paper.\par
Meanwhile, for linguistic feature extraction, we first adopt the GloVe word embeddings \cite{pennington-etal-2014-glove} pre-trained on Common Crawl (840B tokens) to initialize the parameters of embedding layers then a LSTM is employed as the language feature extractor. The LSTM shares parameters to embed each phrase. The corresponding phrases to each image are selected from phrase annotations according to affordance categories. The number of phrases for each image is set to be $4$, and each phrase is embedded to a vector of $C_l =1000$ dimensions.  \par
We train the model using the Adam optimizer \cite{kingma2014adam}. The learning rate is initialized as $2.5 \times 10^{-4}$ with a weight decay of $5\times 10^{-4}$ with gradually decreasing by a polynomial policy with a power of 0.9. The model is trained on an NVIDIA RTX3080 GPU for 100 epochs with a batch size of 1. 

\begin{figure*}[t]
	\centering
		\begin{overpic}[width=1.\linewidth]{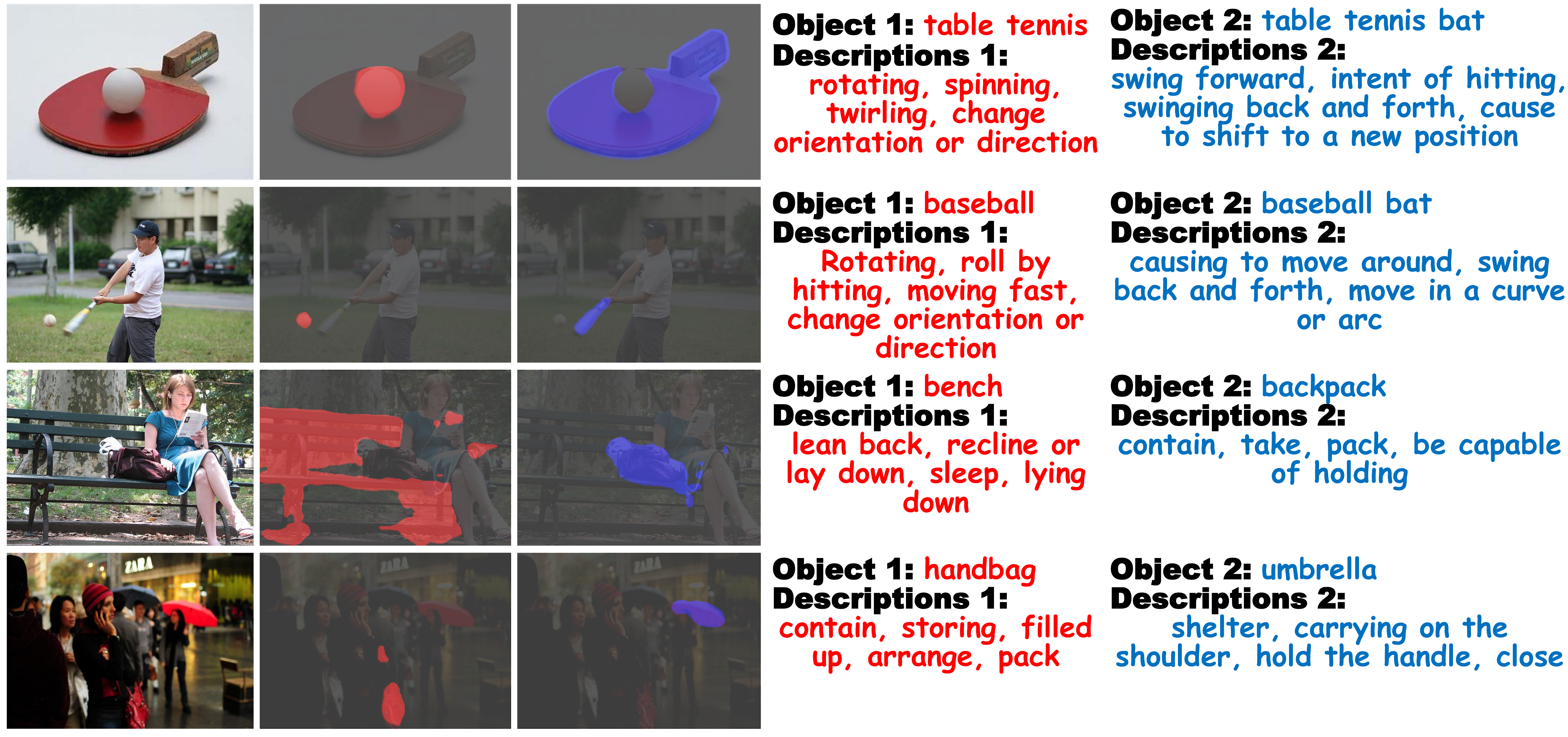}
		\end{overpic}
		\caption{\textbf{Single image $with$ Different descriptions.} When multiple objects with various affordances appear in the same image, our model is expected to highlight the correct object according to the description phrases. The phrases in red indicate the results in the second column, while the blue ones refer to the blue objects in the third column.}
	\label{FIG:sing}
\end{figure*}

\begin{figure*}[t]
	\centering
		\begin{overpic}[width=1.\linewidth]{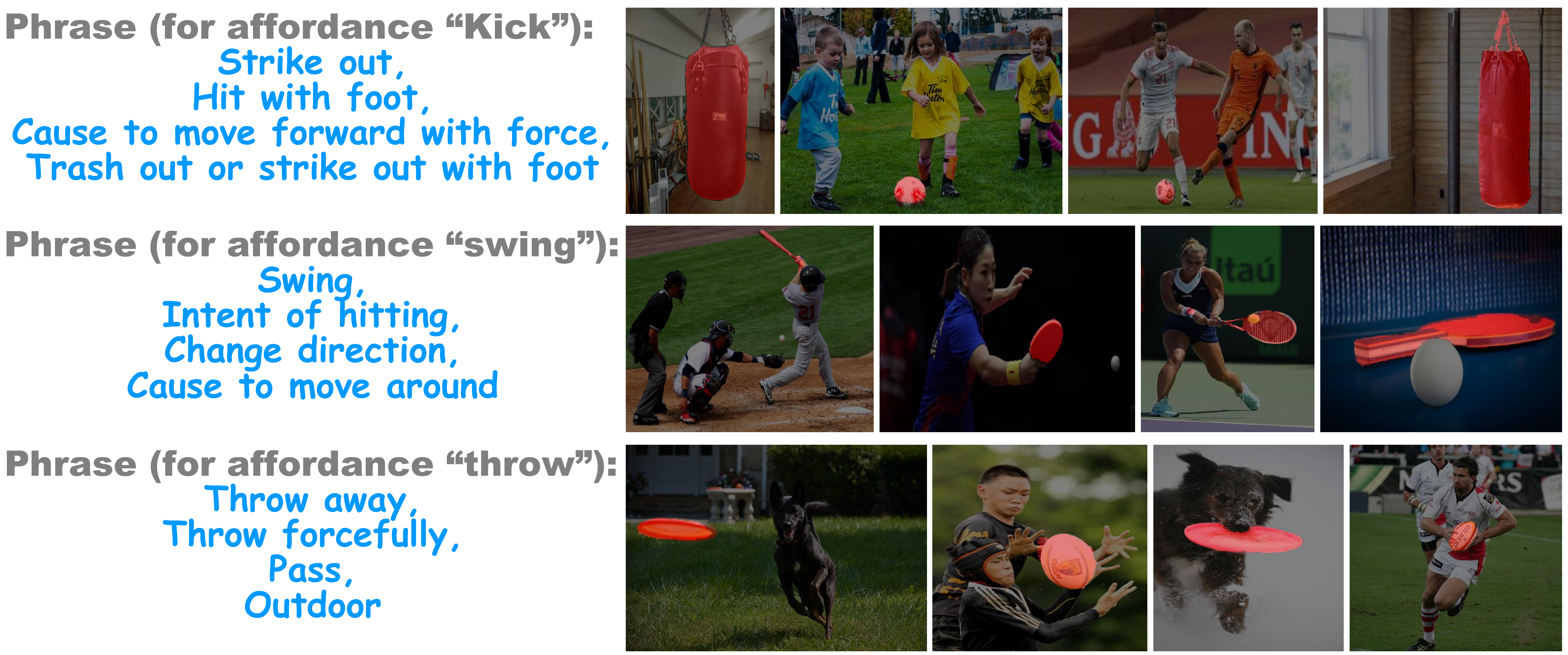}
		\end{overpic}
		\caption{\textbf{Multiple images $with$ Same descriptions.} When we use the same set of affordance phrases, our model is required to segment the objects with the same affordance in multiple images regardless of the appearance variations. The results regions are highlighted in red. Phrases in blue indicate affordance descriptions to the following images in the same row.}
	\label{FIG:multiple}
\end{figure*}

\subsection{Results Analysis}\label{analysis}

We conduct a comprehensive and thorough analysis of the proposed model in this section:

\textbf{\emph{Comparison with other methods:}} \label{results}
We compare our method with several other methods chosen from four fields: semantic segmentation, saliency object detection, affordance detection, and referring segmentation. The results of objective metrics are shown in Table \ref{objective results}. It reveals that our method surpasses all other methods in all metrics. Especially in terms of IoU, $E_{\phi}$ and $F_{\beta}$, our model achieves $2.4\%$, $2.0\%$ and $1.8\%$ performance improvement, respectively. Notably, the table shows that methods involving multi-modalities generally achieve better performance than methods in other fields because of additional language information used. The use of a cyclic interaction mechanism provides our method with better alignments. We also present the subjective visualization results in Fig \ref{FIG:visual}. Our method generates more precise segmentation masks closer to the ground truth than other methods. It indicates that our model can effectively capture the relationship between vision and language. Compared with the multi-modal method CMPC, our approach introduces fewer noises in the background because of more accurate alignment between cross-modal information. For other methods, unexpected objects may be segmented incorrectly because of the absence of necessary language guidance, and some failure cases are shown in the figure.
\par
We also show the IoU scores of all methods in every affordance category in Table. \ref{Table:IoU}. It further demonstrates the superior performance of our proposed model. Our model achieves the best performance in almost all categories except in ``Beat'', ``Bounce'' and ``Brush'' classes. The highest IoU score ($0.766$) is in the affordance class ``Beat'', which only contains the object ``drum'' with similar simple and regular shapes. The lowest IoU score ($0.443$) appears in class ``Brush'' including object ``toothbrush'', which is small and has complicated geometry. It shows that for objects with simple and regular shapes, our model will get higher IoU scores while it is slightly underperforming for small objects.

\textbf{\emph{Single image $with$ Different descriptions:}} \label{SingleToDiff}
To better comprehend the surrounding scene, when there are multiple objects with different affordances in the same image, our model is expected to be able to segment the corresponding regions based on the natural language descriptions. Some examples are shown in Fig. \ref{FIG:sing}. The examples illustrate that our proposed model can align vision and language information correctly even though language changes.

\begin{table}[t]
    \caption{The influence of the \textbf{number of query phrases} on the performance.}
    \centering    
    \renewcommand{\arraystretch}{1.}
    \renewcommand{\tabcolsep}{7.pt}
    \begin{tabular}{c||c|c|c|c|c}
    \hline
    \Xhline{2.\arrayrulewidth}
    \rowcolor{mygray}
    N    & IoU ($\uparrow$) & $F_{\beta}$ ($\uparrow$) & $E_{\phi}$ ($\uparrow$) & CC ($\uparrow$) & MAE ($\downarrow$) \\
    \hline
    \Xhline{2.\arrayrulewidth}
    1    &    $0.532$    &    $0.565$    &  $0.728$    &  $0.671$   & $0.089$   \\
    2    &    $0.563$       &  $0.607$   &  $0.759$    & $0.700$     &  $0.076$ \\
    3    &    $0.580$           & $0.633$    &  $0.788$   & $0.707$ &  $0.068$  \\
    4 (Ours)    &    \bm{$0.593$}   &  \bm{$0.665$}  &     \bm{$0.822$}   & \bm{$0.713$}    &   \bm{$0.061$}\\
    5    &  \underline{$0.585$}      &    \underline{$0.637$}      &    \underline{$0.792$}      & $0.711$  & \underline{$0.067$}    \\
    6    &  \underline{$0.585$}      &    $0.611$   &  $0.760$ &  \underline{$0.712$}    & $0.077$   \\
    \hline
    \Xhline{2.\arrayrulewidth}
    \end{tabular}
    \label{tab:num_influence}
\end{table}

\begin{table}[t]
    \caption{The influence of different \textbf{language encoder method}. \textbf{Bold} and \underline{underline} indicate the best and the second-best scores, respectively}
    \centering    
    \renewcommand{\arraystretch}{1.}
    \renewcommand{\tabcolsep}{6.pt}
    \begin{tabular}{c||c|c|c|c|c}
    \hline
    \Xhline{2.\arrayrulewidth}
    \rowcolor{mygray}
    N    & IoU ($\uparrow$) & $F_{\beta}$ ($\uparrow$) & $E_{\phi}$ ($\uparrow$) & CC ($\uparrow$) & MAE ($\downarrow$) \\
    \hline
    \Xhline{2.\arrayrulewidth}
    LSTM \cite{6795963}   &    \bm{$0.593$}    &    \bm{$0.665$}    &  \bm{$0.822$}    &  \bm{$0.713$}   & \bm{$0.061$}   \\
    BERT \cite{devlin2018bert}  &  \underline{$0.513$}   & \underline{$0.604$}    & \underline{$0.783$}  & \underline{$0.633$}   &   \underline{$0.075$}    \\
   ELMo \cite{Peters2018DeepCW} &  $0.498$   & $0.580$  & $0.775$& $0.631$ &  $0.079$ \\
    \hline
    \Xhline{2.\arrayrulewidth}
    \end{tabular}
    \label{tab:languge_encoder}
\end{table}

\begin{table}[t]
    \caption{The influence of \textbf{the number of cyclic times} which is denoted as $N$ in the first column. In this paper, we only employ CIM module once which can be regarded as a baseline. \textbf{Bold} and \underline{underline} indicate the best and the second-best scores, respectively}
    \centering    
    \renewcommand{\arraystretch}{1.}
    \renewcommand{\tabcolsep}{6.pt}
    \begin{tabular}{c||c|c|c|c|c}
    \hline
    \Xhline{2.\arrayrulewidth}
    \rowcolor{mygray}
    N    & IoU ($\uparrow$) & $F_{\beta}$ ($\uparrow$) & $E_{\phi}$ ($\uparrow$) & CC ($\uparrow$) & MAE ($\downarrow$) \\
    \hline
    \Xhline{2.\arrayrulewidth}
    1 (Baseline)   &    \bm{$0.593$}    &    \underline{$0.665$}    &  \underline{$0.822$}    &  \bm{$0.713$}   & \underline{$0.061$}   \\
    2    &    \underline{$0.588$}   &  \bm{$0.670$}   &  \bm{$0.830$}    & $0.682$     &  \bm{$0.059$} \\
    3    &    $0.572$  & $0.633$    &  $0.788$   & \underline{$0.707$} &  $0.068$ \\
    \hline
    \Xhline{2.\arrayrulewidth}
    \end{tabular}
    \label{tab:num_cim}
\end{table}

\textbf{\emph{Multiple images $with$ Same descriptions:}} \label{DiffToSingle}
In practice, multiple objects may have the same affordance, although with significant appearance variations in terms of color, shape, and texture. Therefore, our model is expected to identify corresponding objects in different images regardless of these variations. Some examples are illustrated in Fig. \ref{FIG:multiple}. From the examples, we can find that for the same set of phrases, corresponding referred objects described by the phrases could be highlighted correctly, which proves that our model can cope with the multiplicity property of affordance.

\subsection{Ablation Study} \label{ablation}

In this section, we conduct ablation study to investigate the effect of different modules and hyper-parameter settings. We consider the following factors: the number of input phrases, the language encoder method, and cyclic times of CIM module. \par

\textbf{\emph{Number of Text Phrases:}}
To explore the influence of the number of input phrases, we set the phrase number to be $N=1,2,3,4,5,6$, respectively. The results are shown in Table \ref{tab:num_influence}. The results illustrate that the phrase number influences all five metrics. It suggests that taking four phrases as input makes the model capture information more effectively. Contrary to intuition, performance does not improve with the number of phrases increasing after the number getting four. Our model may be reached a bottleneck after that point.

\textbf{\emph{Language Encoder Method:}} We consider to explore the effect of different language encoder methods. We replace LSTM with two different popular pre-trained language models BERT \cite{devlin2018bert} and ELMo \cite{Peters2018DeepCW}. The results are illustrates in Table \ref{tab:languge_encoder}. It is shown that LSTM outperforms the other two language encoder methods in this task. The possible reason is that the latter two pre-trained language models are more suitable for long sentences because of rich text context. However, the text descriptions of affordances in the proposed PAD-L are all short phrases that may limit their capabilities.

\textbf{\emph{Number of Cycles:}} The semantic consistency is enhanced in a cyclic and bilateral manner. To investigate the effect of the number of cycles, we repeat CIM module several times. The results are shown in Table. \ref{tab:num_cim}. It is demonstrated that more cyclic times do not necessarily lead to better performance. We set cycling once as the baseline. When CIM repeats twice, the performance outperforms the baseline in several metrics. However, the performance is not as good as the baseline when cycle three times. Our model may get stuck in over-fitting as the number of cycle increases.


\section{Conclusion and Discussion} \label{conclude}

In this paper, we propose a novel phrase-based affordance detection task. At first, based on the previously proposed PAD dataset, we annotate the affordance categories using short phrases to construct a new multi-modal dataset, PAD-Language (PAD-L). Then to align text features and vision features better, we adopt a novel cyclic and bilateral mechanism to cope with the problem caused by the inherent multiplicity property of affordance. Specifically, we design a Cyclic Bilateral Consistency Enhancement Network (CBCE-Net), which consists of three main modules: Vision guide Language Module (VLM), Language guide Vision Module (LVM), and Cyclic Interaction Module (CIM) to improve feature representations cyclically and bilaterally. Compared with nine relevant methods, our model achieves the best results in terms of all five evaluation metrics. \par
Our approach also has some limitations. At first, our method may not achieve satisfactory results in complicated scenes containing many objects. To improve our method, we could adopt a locate-then-segment framework to locate objects \cite{wu2021background} then generate the mask. Secondly, Our approach aims at detecting all possible objects in the image and cannot detect the one that best fits the intention. We can introduce a sorting mechanism to segment the objects that best match the actual scene. \par
In the future, based on PAD-L, more works could be done to explore the combination of multi-modal applications and affordance. For instance, exploring affordance detection in videos with natural language instructions would be a promising topic.

\bibliographystyle{IEEEtran}
\bibliography{IEEEtranbib}

\clearpage

\setcounter{page}{1}

\title{Phrase-Based Affordance Detection \\ via Cyclic Bilateral Interaction}

\author{Liangsheng Lu, Wei Zhai, Hongchen Luo, Yu Kang, 
\IEEEmembership{Senior Member, IEEE} and Yang Cao, \IEEEmembership{Member, IEEE}

\thanks{L. Lu, W. Zhai, H. Luo, Y. Kang, Y. Cao are at the University of Science and Technology of China, Anhui, China. (email: \{lulsheng, wzhai056, lhc12\}@mail.ustc.edu.cn, \{kangduyu, forrest\}@ustc.edu.cn).}}

\maketitle
\setcounter{page}{1}

\section{Appendix}

We shows the details of our evaluation metrics in Section \ref{setting} and comparison methods in Section \ref{compare}.
Also, more examples of phrases annotations are shown in Table \ref{examples}

\begin{table*}[h!] \label{examples}
\centering
  \renewcommand{\arraystretch}{1.}
    \renewcommand{\tabcolsep}{2.pt}
  \caption{More examples of phrases descriptions about affordances. \textbf{PA}, \textbf{F}, \textbf{AF}, \textbf{E} denote \textbf{P}otential \textbf{A}ctions, \textbf{F}unction, \textbf{A}ppearance \textbf{F}eature and \textbf{E}nvironment, respectively in the table. Note that we only show the original form of the corresponding verbs.} 
  \label{Affordance Description table}
  \begin{tabular}{c||c|c}
  \Xhline{2.\arrayrulewidth}
    \hline
  \textbf{Affordance Class} & \textbf{Object Class} & \textbf{Phrase Descriptions Examples} \\
  \hline
  \Xhline{2.\arrayrulewidth}
  \rowcolor{mygray}
    \textbf{\normalsize{Kick}}     &    \multicolumn{1}{m{3.7cm}|}{\small{soccer ball, punching bag}}     & \multicolumn{1}{m{11cm}}{
    \textbf{PA}: move fast, trash out or strike, make a motion with feet or fist toward an object, strike out with feet, punt, physical strike, ...  
    \qquad                  
    \textbf{E}: outdoor activities (soccer ball) }  \\
 
     \textbf{\normalsize{Sit}}     &    \multicolumn{1}{m{3.7cm}|}{\small{bench, sofa, stool, wheelchair}}         & \multicolumn{1}{m{11cm}}{
    \textbf{PA}: sit, sit down, seat, lounge, recline, be seated, sit in, lean back, lean over, lean against, ... 
    \quad 
    \textbf{F}: rest, take a rest, sleep, nap, take a break, have a rest, give feet a rest... 
    }  \\
    \rowcolor{mygray}
     \textbf{\normalsize{Throw}}    &    \multicolumn{1}{m{3.7cm}|}{\small{frisbee, rugby ball}}      &  \multicolumn{1}{m{11cm}}{
    \textbf{PA}: 
    throw, deliver, pass, toss, toss using hands, throw away, throw forcefully, cast, ... 
    \quad 
    \textbf{E}: outdoor, out-of-doors, 
    }          \\

    \textbf{\normalsize{Shelter}}      &    \multicolumn{1}{m{3.7cm}|}{\small{umbrella}}    &  \multicolumn{1}{m{11cm}}{
    \textbf{PA}: shelter, raise, lift, move up, carry, take, grip handle, take, ... \quad
    \textbf{F}: cover for, protect, shade, shield \quad 
    \textbf{E}: in the sun, in the rain, outdoor, \quad 
    \textbf{AF}: circular cover
    }   \\
    \rowcolor{mygray}
     \textbf{\normalsize{Beat}}   &   \multicolumn{1}{m{3.7cm}|}{\small{drum}}      &  \multicolumn{1}{m{11cm}}{
    \textbf{PA}: beat,  strike, hit, strike rapidly, hit in rhythm, pulse, beat in rhythm, clout, punch, pound, ... \quad
    \textbf{F}: play, sound, create sound, make sound, produce sound
    }\\

     \textbf{\normalsize{Hit}}   &   \multicolumn{1}{m{3.7cm}|}{\small{axe, hammer}}        & \multicolumn{1}{m{11cm}}{
    \textbf{PA}: hit, deliver an impulsive fore by striking, strike, can be lifted \quad
    \textbf{F}: hit, chop, split, cut, cleave \quad
    \textbf{AF}: sharp blade, knife-edged \quad
    \textbf{E}: usually appears along with wood
    }   \\
    \rowcolor{mygray}
     \textbf{\normalsize{Cut}}   &   \multicolumn{1}{m{3.7cm}|}{\small{knife, scissors}}      & \multicolumn{1}{m{11cm}}{
    \textbf{PA}: cut, hold, use, sharpen, grasp, raise, slash, pull into, hold the handle, ... \quad
    \textbf{F}: separate, slice, chop, divide, part, trim, ... \quad
    \textbf{AF}: sharp edge,  usually made of metal \quad
    } \\

     \textbf{\normalsize{Lie}}  &   \multicolumn{1}{m{3.7cm}|}{\small{baby bed, bench, sofa}}  & \multicolumn{1}{m{11cm}}{
    \textbf{PA}: lie, lie down, sit down, recline or lay down, lean back, lean over, be recumbent, sit back, lie on the side, prostrate, lean, ... \quad
    \textbf{F}: take a break, sleep, rest, repose, ... \quad
    }      \\
    
    \rowcolor{mygray}
     \textbf{\normalsize{Lift}}     &   \multicolumn{1}{m{3.7cm}|}{\small{dumbbell}}    &  \multicolumn{1}{m{11cm}}{
    \textbf{PA}: lift, lift up, raise, grab, put down, pick up, take down, push, hold up, uplift, cause to raise, hold high,  \quad
    \textbf{F}: exercise, used for exercise of muscle-building \quad
    \textbf{E}: indoor exercise 
    }        \\

     \textbf{\normalsize{Pick up}}  &    \multicolumn{1}{m{3.7cm}|}{\small{chopsticks}}    &  \multicolumn{1}{m{11cm}}{
    \textbf{PA}: take and lift upward, hold, grasp, move up and down, hold and lift \quad 
    \textbf{F}: pass food, kitchen utensil \quad
    \textbf{E}: usually appears in kitchen or dining table\quad
    \textbf{AF}: usually are made of wood\quad
    }            \\
    \rowcolor{mygray}
     \textbf{\normalsize{Rolling}}   &    \multicolumn{1}{m{3.7cm}|}{\small{baseball, croquet ball, golf ball, table tennis ball, tennis ball}}      &  \multicolumn{1}{m{11cm}}{
    \textbf{PA}: rolling, move, can roll, move by rotating, roll over, rotate rapidly, turn round and round, rotate, move fast, spin, whirl, move around an axis or a center, cycle, revolve, change orientation or direction, twirl revolve \quad
    \textbf{AF}: spherical
    }                   \\

     \textbf{\normalsize{Mix}} & \multicolumn{1}{m{3.7cm}|}{\small{chopsticks, spoon, whisk}} &  \multicolumn{1}{m{11cm}}{
    \textbf{PA}: mix, blend, mix together, fuse, grasp, hold, merge, move circularly, move around, agitate, ... \quad
    \textbf{F}: kitchen tools \quad
    \textbf{E}: usually appears in kitchen or dining table,
    } \\
    \rowcolor{mygray}
     \textbf{\normalsize{Jump}}  & \multicolumn{1}{m{3.7cm}|}{\small{skateboard, skis, snowboard, surfboard}} & \multicolumn{1}{m{11cm}}{
    \textbf{PA}: jump, turn at high speed, move forward, move fast, travel fast, perform a leap, accelerate, make a turn, speed, turn left, turn right, make a turn, speed up, ... \quad
    \textbf{E}: outdoor activities. 
    } \\

      \textbf{\normalsize{Fork}}     &    \multicolumn{1}{m{3.7cm}|}{\small{fork}}   & \multicolumn{1}{m{11cm}}{
    \textbf{PA}: fork, fork up, move up and down, hold handle \quad
    \textbf{F}: pass food, pick up food, used for cook, lift food\quad
    \textbf{E}: appears in kitchen or dining table, used with knife\quad
    }     \\
    \rowcolor{mygray}
      \textbf{\normalsize{Scoop}}      &  \multicolumn{1}{m{3.7cm}|}{\small{spatula, spoon}}        & \multicolumn{1}{m{11cm}}{
    \textbf{PA}: scoop, scoop out, scoop up, take up, ladle out, hold the handle, grasp the handle, lade, take out or up \quad
    \textbf{E}: appears in the kitchen or the dining table \quad
    \textbf{AF}: concave shape
    }        \\

     \textbf{\normalsize{Swing}}   &  \multicolumn{1}{m{3.7cm}|}{\small{baseball bat, table tennis bat, tennis racket}}   & \multicolumn{1}{m{11cm}}{
    \textbf{PA}: swing, change location by moving back and forth, change direction, cause to move around, swing back, swing forward, swing back and forth, try to hit something \quad
    }              \\
    \rowcolor{mygray}
     \textbf{\normalsize{Take photo}}   &  \multicolumn{1}{m{3.7cm}|}{\small{camera}}       & \multicolumn{1}{m{11cm}}{
    \textbf{PA}: shoot, take a shot of, target to, adjust, put in front of eyes, aim at, raise up to eyes, bring up to eyes, snap, keep \quad
    \textbf{F}: take a photo of, get pictures of, capture in a photo
    }              \\

      \textbf{\normalsize{Bounce}}   &  \multicolumn{1}{m{3.7cm}|}{\small{basketball}}   & \multicolumn{1}{m{11cm}}{
    \textbf{PA}: bounce, spring back, move up and down, rebound, bounce back, move quickly back and forth, pass, bounce against, ... \quad
    \textbf{AF}: bouncy, spherical, rubber or synthetic material, ... \quad
    \textbf{E}: usually in door, team sport
    }      \\
    \rowcolor{mygray}
    
     \textbf{\normalsize{Contain-1}}   &  \multicolumn{1}{m{3.7cm}|}{\small{backpack, gift box, handbag, purse, suitcase}} & \multicolumn{1}{m{11cm}}{
    \textbf{PA}: contain, take, hold, have within, pack, pack into, place within, hold in, fill up, load up, make full \quad
    \textbf{F}: hold household items, hold inside, store, be capable of holding
    }   \\

     \textbf{\normalsize{Contain-2}}    &  \multicolumn{1}{m{3.7cm}|}{\small{beaker, beer bottle, bowl, cup or mug, milk can, pitcher, soap dispenser, vase, watering can}}     & \multicolumn{1}{m{11cm}}{
    \textbf{PA}: contain, pour, hold, pour in, pour out, decant, flow, store, keep, hold in, carry, bear, have within, include, take, pour off, hold in hands, dribble, spill, ... \quad
    \textbf{AF}: depression in the middle, open-top container,  contain liquid, liquid container, ...
    }              \\
    
    \rowcolor{mygray}
      \textbf{\normalsize{Contain-3}}   &  \multicolumn{1}{m{3.7cm}|}{\small{bowl, frying pan}}    & \multicolumn{1}{m{11cm}}{
    \textbf{PA}: contain, store, hold in both hands up\quad 
    \textbf{F}: prepare for food, hold and store food \quad
    \textbf{AF}: the center is depressed, depression in the middle \quad 
    \textbf{E}: usually appears in kitchen or dining table
    }  \\

     \textbf{\normalsize{Play-1}}       &  \multicolumn{1}{m{3.7cm}|}{\small{cell, erhu fiddle, viola, violin}}       & \multicolumn{1}{m{11cm}}{
    \textbf{PA}: play, bow, fiddle, chord, press strings, squeeze the bow, move bow across strings, grip the bow, ... \quad
    \textbf{F}: make sound, make music, produce sound, stringed instruments, ... 
    }\\
    
    \rowcolor{mygray} 
      \textbf{\normalsize{Play-2}}        &  \multicolumn{1}{m{3.7cm}|}{\small{banjo, guitar, harp, pipa}}       & \multicolumn{1}{m{11cm}}{
    \textbf{PA}:  play, carry, move fingers up and down, pluck fingers, press the string, perform, pull slightly but sharply, ... \quad
    \textbf{F}: make sound, make music, produce sound, stringed musical instrument, ...
    }  \\
    
     \textbf{\normalsize{Play-3}}       &  \multicolumn{1}{m{3.7cm}|}{\small{accordion, piano}}       & \multicolumn{1}{m{11cm}}{
    \textbf{PA}: play, tune, press keys, move fingers, touch, manipulate, squeeze, ... \quad
    \textbf{F}: make sound, produce music, make music, ...
    }    \\
    \rowcolor{mygray}
     \textbf{\normalsize{Play-4}}       &  \multicolumn{1}{m{3.7cm}|}{\small{flute, frenchhorn, harmonica, trumpet}}       & \multicolumn{1}{m{11cm}}{
    \textbf{PA}: play, tune, hold, blow air into the instrument, raise to lip, perform, push aside mouth, lift to lip, carry, blow through mouth, carry, wind, ... \quad
    \textbf{F}: make sound, make music, produce sound
    }    \\
     \textbf{\normalsize{Ride}}       &  \multicolumn{1}{m{3.7cm}|}{\normalsize{bicycle, motorbike}}       & \multicolumn{1}{m{11cm}}{
    \textbf{PA}: ride, push down with foot, pedal, turn left, move rapidly, pull, control motion, slow down, stop, ... \quad
    \textbf{F}: travel, change location, travel fast, ... \quad
    \textbf{E}: outdoor
    }    \\
    \rowcolor{mygray}
     \textbf{\normalsize{Brush}}       &  \multicolumn{1}{m{3.7cm}|}{\normalsize{toothbrush}}       & \multicolumn{1}{m{11cm}}{
    \textbf{PA}: brush, grasp the handle, hold handle, touch lightly and briefly, ... \quad
    \textbf{F}: clean, sweep, rub, sweep across or over, wash, clean tooth, ... \quad
    \textbf{AF}: head attached to a handle, a head of tightly clustered bristles, ... \quad
    \textbf{E}: often appears beside a sink within the kitchen or bathroom, ...
    }    \\
     \textbf{\normalsize{Roll dough}}       &  \multicolumn{1}{m{3.7cm}|}{\normalsize{rolling pin}}       & \multicolumn{1}{m{11cm}}{
    \textbf{PA}: roll, press, roll the rod across the dough, grasp the handle, shape, shape by rolling, squeeze, shape by rolling, exert a force with a heavy weight, ... \quad
    \textbf{AF}: cylindrical, ... \quad
    \textbf{E}: appear in the kitchen, ... \quad
    \textbf{F}: food preparation utensil, kitchen stuff, ... \quad 
    }    \\
    \rowcolor{mygray}
     \textbf{\normalsize{Wear-1}}       &  \multicolumn{1}{m{3.7cm}|}{\normalsize{hat, helmet}}       & \multicolumn{1}{m{11cm}}{
    \textbf{PA}: wear, put on, take off, dress, be dressed in, be clothed in, carry, get dressed, hold, keep, raise, cover, have on, ... \quad
    \textbf{F}: decorate, protect against, shelter from the sun, head covering, have on, used for warmth, ...
    }    \\
     \textbf{\normalsize{Wear-2}}       &  \multicolumn{1}{m{3.7cm}|}{\normalsize{glasses}}       & \multicolumn{1}{m{11cm}}{
    \textbf{PA}: wear, wear on face, take off, put off, put on, raise, get, ... \quad
    \textbf{AF}: two pieces of glasses, \quad
    \textbf{F}: improve vision, protect eyes, used for decoration, ...
    }    \\
    \rowcolor{mygray}
     \textbf{\normalsize{Look Out}}       &  \multicolumn{1}{m{3.7cm}|}{\normalsize{binoculars}}       & \multicolumn{1}{m{11cm}}{
    \textbf{PA}: look out, adjust, hold in hands, target to, focus, look at, set the focus, put in front of eyes, aim at, zoom, bring up to eyes, turn the focus wheel, align with view, adjust, ... \quad
    \textbf{F}: see clearly  \quad
    \textbf{E}: outdoor \quad
    \textbf{AF}: two lens, two telescopes mounted side by side
    }    \\
    
    \hline
    \Xhline{2.\arrayrulewidth}
    \end{tabular}
\end{table*}

\subsection{Benchmark Setting} \label{setting}

We choose five broadly used metrics to comprehensively evaluate the performance of different methods, \ie, Intersection over Union (IoU), F-measure ($F_{\beta}$), E-measure ($E_{\phi}$), Pearson's Correlation Coefficient (CC), and Mean Absolute Error (MAE). We introduce them briefly as following:

\begin{itemize}[leftmargin=*]
    \item
    \textbf{Intersection of Union (IoU) \cite{long2015fully}}: IoU is a common pixel level evaluation metric to measure the  overlap between predicted mask and ground truth mask. It is defined as the ratio of the area of overlap and the area of union.
    \item 
    \textbf{F-measure ($F_{\beta}$) \cite{arbelaez2010contour}}: $F_{\beta}$ is a widely used metric which simultaneously considers both recall $R$ and precision $P$, where $P$ is the number of true positive results divided by the number of all positive results and $R$ is the number of true positive results divided by the number of all samples that should have been identified as positive. 
    
    \item 
    \textbf{E-measure ($E_{\phi}$) \cite{fan2018enhanced}}: $E_{\phi}$ is a measurement which jointly utilizes local and global information to evaluate the difference between the ground-truth and predicted mask.
    
    \item 
    \textbf{Pearson's Correlation Coefficient (CC) \cite{le2007predicting}}: CC is broadly applied to measure the linear correlation between two variables. In this paper, we employ CC to measure the relevance of the predicted map and the ground truth.
    
    \item
    \textbf{Mean Absolute Error (MAE) \cite{perazzi2012saliency}}: MAE measures the average over the absolute differences of the normalized predicted map and the ground-truth mask.
\end{itemize}

The evaluation code can be found at \url{https://github.com/lhc1224/OSAD_Net/tree/main/PyMetrics}.

\subsection{Comparison Methods} \label{compare}

To illustrate the superiority of our model, we compare several different kinds of methods, which involve \textbf{two} \textcolor[rgb]{0.8,0.0,0.3}{\textbf{Salient Detection} models (BASNet, CPD)}, \textbf{two} \textcolor[rgb]{0.99,0.5,0.0}{\textbf{Affordance Detection} models (OSAD-Net, OAFFD)}, \textbf{two} \textcolor[rgb]{0.4,0.0,0.99}{\textbf{Semantic Segmentation} models (PSPNet, DeepLabV3+)}, and \textbf{three} \textcolor[rgb]{0.1,0.8,0.1}{\textbf{Referring Segmentation} models (CMSA, BRINet, CMPC)}.

\begin{itemize}[leftmargin=*]
    \item 
    \textcolor[rgb]{0.8,0.0,0.3}{\textbf{BASNet}} \cite{Qin_2019_CVPR}: \textbf{B}oundary-\textbf{A}ware \textbf{S}egmentation \textbf{N}etwork consists of a predict-refine architecture and a hybrid loss. The predict-refine architecture consists of a encoder-decoder network and a refinement module to predict and refine the segmentation probability map respectively.
    
    \item
    \textcolor[rgb]{0.8,0.0,0.3}{\textbf{CPD}} \cite{Wu_2019_CVPR}: \textbf{C}ascaded \textbf{P}artial \textbf{D}ecoder (CPD) framework leverages partial decoder to discard large resolution features in shallower layers and integrate features of deeper layers to generate more precise saliency map. 
    
    \item 
    \textcolor[rgb]{0.99,0.5,0.0}{\textbf{OSAD-Net}} \cite{Oneluo}: \textbf{O}ne \textbf{S}hot \textbf{A}ffordance \textbf{D}etection \textbf{N}etwork first learns the intentions of the human actions and then transfers it to query images to segment objects with the same affordance by collaborative learning. 
    
    \item 
    \textcolor[rgb]{0.99,0.5,0.0}{\textbf{OAFFD}} \cite{zhao2020object}: OAFFD-Net mainly combines CoordConv and ASPP to refine the feature maps, and designs a relationship-aware module to explore the relationships between objects and affordance.
    
    \item 
    \textcolor[rgb]{0.4,0.0,0.99}{\textbf{PSPNet}} \cite{zhao2017pspnet}: \textbf{P}yramid \textbf{S}cene \textbf{P}arsing \textbf{Net}work utilizes a pyramid parsing module to exploit global context information. Thus the local and global clues are used together to improve the performance in semantic segmentation task.
    
    \item 
    \textcolor[rgb]{0.4,0.0,0.99}{\textbf{DeepLabV3+}} \cite{chen2017rethinking}: \textbf{DeepLabV3+} applies the depthwise separable convolution to an \textbf{A}trous \textbf{S}patial \textbf{P}yramid \textbf{P}ooling (ASPP) model to encode multi-scale context information at multiple filter rates and multiple fields-of-view. 
    
    \item 
    \textcolor[rgb]{0.1,0.8,0.1}{\textbf{CMSA}} \cite{ye2021referring}: \textbf{C}ross-\textbf{M}odal \textbf{S}elf-\textbf{A}ttention module is able to adaptively focus on the important words in the given language expression and region in the corresponding image by utilizing self-attention mechanism.
    
    \item 
    \textcolor[rgb]{0.1,0.8,0.1}{\textbf{BRINet}} \cite{hu2020bi}: \textbf{B}i-directional \textbf{R}elationship \textbf{I}nferring \textbf{N}etwork designs two kinds of attention mechanism from vision to language and language to vision to learn the bi-directional relationship between language and visual modalities. 

    \item 
    \textcolor[rgb]{0.1,0.8,0.1}{\textbf{CMPC}} \cite{liu2021cross}: \textbf{C}ross-\textbf{M}odal \textbf{P}rogressive \textbf{C}omprehension scheme first perceives all related entities utilizing entity and attribute words while the rest relational words are adopted to highlight the target entities by spatial graph reasoning.
    
\end{itemize}

\end{document}